\documentclass{article}
\pdfoutput=1
\PassOptionsToPackage{numbers, compress}{natbib}
\usepackage[preprint]{nips_2018}

\usepackage{times}
\usepackage[utf8]{inputenc} % allow utf-8 input
\usepackage[T1]{fontenc}    % use 8-bit T1 fonts
\usepackage{hyperref}       % hyperlinks
\usepackage{url}            % simple URL typesetting
\usepackage{booktabs}       % professional-quality tables
\usepackage{amsmath}
\usepackage{mathtools}
\usepackage{amsfonts}       % blackboard math symbols
\usepackage{nicefrac}       % compact symbols for 1/2, etc.
\usepackage{microtype}      % microtypography
\usepackage{color}
\usepackage{subcaption}
\usepackage{comment}
\usepackage{xcolor}
\usepackage{cancel}
\usepackage{sidecap}
\sidecaptionvpos{figure}{t}
\usepackage{graphicx}
\usepackage{wrapfig}

\title{Learning Plannable Representations with Causal InfoGAN}

\author{
  Thanard Kurutach\thanks{equal contribution, correspondence to \texttt{\{thanard.kurutach, avivt\}@berkeley.edu}} \\
  UC Berkeley \\
  \And
  Aviv Tamar\footnotemark[1] \\
  UC Berkeley \\
  \And
  Ge Yang \\
  University of Chicago \\
  \AND
  Stuart Russell \\
  UC Berkeley \\
  \And
  Pieter Abbeel \\
  UC Berkeley \\
}

\DeclareMathOperator*{\argmin}{arg\,min}
\DeclareMathOperator*{\argmax}{arg\,max}

\newcommand{\data}{\mathcal{D}}
\newcommand{\model}{\mathcal{M}}

\newcommand{\States}{S}

\newcommand{\PD}{P_{\text{data}}}
\newcommand{\PN}{P_{\text{noise}}}

\newtheorem{problem}{Problem}

\definecolor{ge}{HTML}{23AAFF}

\definecolor{thanard}{HTML}{23AAFF}

\def\viewcomments{0}

\ifx\viewcomments\undefined
  \newcommand{\AT}[1]{\iffalse{[AT: #1]}\fi}
  \newcommand{\TK}[1]{\iffalse{[TK: #1]}\fi}
  \newcommand{\GY}[1]{\iffalse{[GY: #1]}\fi}
  \newcommand{\TODO}[1]{\iffalse{[TODO: #1]}\fi}
  \newcommand{\DONE}[1]{\iffalse{[DONE: #1]}\fi}
\else
\newcommand{\AT}[1]{\textbf{\textcolor{magenta}{[AT: #1]}}}
\newcommand{\TK}[1]{\textbf{\textcolor{blue}{[TK: #1]}}}
\newcommand{\GY}[1]{\textbf{\textcolor{ge}{[Ge Y.: #1]}}}
\newcommand{\TODO}[1]{\textbf{\textcolor{red}{[TODO: #1]}}}
\newcommand{\DONE}[1]{\textbf{\textcolor{ge}{[DONE: #1]}}}
\fi

\newcommand{\norm}[1]{\left\lVert#1\right\rVert}

\begin{document}

\maketitle

\begin{abstract}
  In recent years, deep generative models have been shown to `imagine' convincing high-dimensional observations such as images, audio, and even video, learning directly from raw data. In this work, we ask how to imagine \emph{goal-directed visual plans} -- a plausible sequence of observations that transition a dynamical system from its current configuration to a desired goal state, which can later be used as a reference trajectory for control.
  We focus on systems with high-dimensional observations, such as images, and propose an approach that naturally combines representation learning and planning.
  Our framework learns a generative model of \emph{sequential observations}, where the generative process is induced by a transition in a low-dimensional \emph{planning model}, and an additional noise. By maximizing the mutual information between the generated observations and the transition in the planning model, we obtain a low-dimensional representation that best explains the causal nature of the data. We structure the planning model to be compatible with efficient planning algorithms, and we propose several such models based on either discrete or continuous states. Finally, to generate a visual plan, we project the current and goal observations onto their respective states in the planning model, plan a trajectory, and then use the generative model to transform the trajectory to a sequence of observations.
  We demonstrate our method on imagining plausible visual plans of rope manipulation. 
\end{abstract}
% \begin{abstract}
%   We study the problem of long-horizon planning in dynamical systems with high-dimensional observations, such as images. We propose a framework for learning low-dimensional and structured representations of observations from a dynamical system, which can be used for planning with conventional AI 
%   planning algorithms. Our framework learns a generative model of sequential observations, where the generative process is induced by a transition in a low-dimensional \emph{planning model}, and an additional noise. By maximizing the mutual information between the generated observations and the transition in the planning model, we obtain a low-dimensional representation that best explains the causal nature of the data. Our framework can be used with planning models that comply with modern search-based planning algorithms, such as discrete binary representations. We show that our method can perform plausible planning of visual rope manipulation. 
% \end{abstract}
  % Ge: why AI planning algorithms? would something like "with conventional planning algorithms" be more general? Sentences afterward all point to modern search planning algos.
% \vspace{-1em}
\section{Introduction}
% \vspace{-1em}
For future robots to perform general tasks in unstructured environments such as homes or hospitals, 
% \TK{Add motivating examples like cooking}
they must be able to reason about their domain and plan their actions accordingly. In AI literature, this general problem has been investigated under two main paradigms -- automated planning and scheduling~\citep{russel2010AI} (henceforth, AI planning) and reinforcement learning~\citep{sutton1998reinforcement} (RL).
  
Classical work in AI planning has drawn on the remarkable capability of humans to perform long-term reasoning and planning by using abstract representations of the world. For example, humans might think of "cup on table" as a state rather than detailed coordinates or a precise image of such a scene. Interestingly, powerful classical planners exist that can reason very effectively with these kinds of representations, as demonstrated by results in the International Planning Competition~\cite{vallati20152014}.
%for example, A$^*$~\cite{hart1968formal}, or the STRIPS planning line of work~\cite{fikes1971strips}. 
% In AI planning, given a description of the environment dynamics, and initial and goal states, a search is performed for a sequence of actions that lead from the initial state to the goal. A fundamental step in AI planning is choosing a representation for the environment model, and mapping the agent's observation onto this representation. 
%TODO say something about successes of planning using good representations, PDDL, etc.
However, such logical representations of the world can be difficult to specify correctly. As an example, consider designing a logical representation for the state of a deformable object such as a rope. Moreover, logical representations that are not grounded {\em a priori} in real-world observation require a perception module that can identify, for example, exactly when the cup is considered "on the table".
% ; (ii) they are a completely isolated logical representation that isn't in any way grounded in the real world. Thus, 
%%AT: Not sure I understand what the isolation means, but I think it relates to perception?
Indeed, most planning successes to date (e.g.,~\cite{nilsson1984shakey, kochenderfer2012next, srivastava2015tractability}) relied on a human-designed state representation, and manually designed the perception of the state from the observation. 

% While this approach led to many successes, such as TODO, in some domains, manually identifying the state can be challenging. For example, consider a robot manipulating a deformable object such as a rope. Representing the state of the rope can be challenging.

In RL, on the other hand, a task is solved directly through trial and error experimentation, guided by a manually provided reward signal. Recent advances in RL using deep neural networks (e.g.,~\cite{mnih2015human,finn2016endtoend}) have shown remarkable success in learning policies that act directly on high-dimensional perceptual inputs, such as raw images. Designing a reward function that depends on high-dimensional observations can be challenging, however, and most recent studies either relied on domains where the reward can be instrumented \cite{mnih2015human,finn2016endtoend,riedmiller2018learning}, or required successful demonstrations as guidance~\cite{finn2016guided,srinivas2018universal}.
Moreover, since RL is guided by the reward to solve a particular task, it does not automatically generalize to different tasks \cite{tamar2016value,kansky17a}. Recent approaches that aim to achieve generalization in RL through learning on a variety of different tasks (e.g.,~\cite{duan2016rl,WangKTSLMBKB16,finn2017model}) are typically not sample-efficient and are limited to relatively simple decision-making problems. 
%% TK: Not only RL doesn't generalize to different tasks, the reward function for a specific task also depends on the state space (not pixel space), which we have no access to in real-world data. 
%%AT: good point, added.

In principle, model-based RL approaches can solve the generalization problem by learning a model of the environment dynamics and planning in that model. However, applying model-based RL to domains with high-dimensional observations has been challenging \cite{watter2015embed,finn2016deep_spatial,finn2017deep}.
Deep learning approaches to learning dynamics models (e.g., action-conditional video prediction models~\citep{oh2015action,agrawal2016learning,finn2017deep}) tend to get bogged down in pixel-level detail, tend to be computationally expensive, and are far from accurate over longer time scales. Moreover, the representations learned using such approaches 
% Recent work in self-supervised video prediction for robot manipulation~\citep{agrawal2016learning,finn2017deep} has shown success in learning representations for image observations of dynamical systems. However, the representations learned in those works 
are typically unstructured, high-dimensional continuous vectors, which cannot be used in efficient planning algorithms. Indeed, prior work has used myopic or random-search-based action selection for planning~\citep{agrawal2016learning,finn2017deep}, which can be effective for planning simple skills such as pushing an object to a target, but does not scale up to more complex, high-level decision making problems such as laying the table for dinner.
%% TK: Can mention model bias & compounding error problems.
%% AT: I think the 'far from accurate over longer time scales' sentence captures the essence of it, no?

In this work, we aim to combine the merits of deep learning dynamics models and classical AI planning, and propose a framework for long-term reasoning and planning that is grounded in real-world perception. We present \textbf{Causal InfoGAN}, a method for learning \emph{plannable representations} of dynamical systems with high-dimensional observations such as images. By plannable, we mean representations that are structured in such a way that makes them amenable for efficient search, through AI planning tools. In particular, we focus on discrete and deterministic dynamics models, which can be used with graph search methods, and on  continuous models where planning is done by linear interpolation, though our framework can be generalized to other model types. 

In our framework, a generative adversarial net (GAN; \cite{goodfellow2014generative}) is trained to generate sequential observation pairs from the dynamical system. The generative network (GAN generator) is structured as a deep neural network that takes as input both unstructured random noise and a structured pair of consecutive states from a low-dimensional, parametrized dynamical system termed the \emph{planning model}. The planning model is meant to capture the features that are \emph{most essential for representing the causal properties} in the data, and are therefore important for planning future outcomes. If the planning model is compliant with efficient planning algorithms and is also informative about the high-dimensional observation sequences, then planning using it should be both computationally efficient and also relevant to planning in the actual system we care about. To induce such an informative model, we follow the InfoGAN idea~\citep{chen2016infogan}, and add to the GAN training a loss function that maximizes the mutual information between the observation pairs and the transitions that induced them. %We therefore term our approach \emph{Causal InfoGAN}. 

We train a causal InfoGAN model using data from random exploration in the system. After learning, given an observation of an initial configuration and a goal configuration, we use our model to generate a ``walkthrough'' sequence of feasible observations that lead from the initial state to the goal. We do this by computing a trajectory in the planning model and using the GAN to transform this trajectory into a sequence of observations. This walkthrough, which breaks the long-horizon planning into a sequence of short-horizon skills, can be later used as a guiding signal for executing the task in the real system.

We compare the representations learned in Causal InfoGAN to standard methods for state aggregation on synthetic tasks, and demonstrate that Causal InfoGAN can generate convincing walkthrough sequences for manipulating a rope into a given shape, using real image data collected by Nair et al.~\cite{nair2017combining} of a robot randomly poking the rope. 

%% TK: I like the way we explicitly separate the high-level planning part from the lower-level modules, and explicitly say that we show the first part.

% \vspace{-1em}
\section{Preliminaries and Problem Formulation}\label{s:problem_formulation}
% \vspace{-1em}
In this section we present background material and our problem formulation.

\subsection{Deep Generative Models based on GAN and InfoGAN }

Let $o\sim \PD(o)$ denote observations sampled from a dataset. Deep generative models aim to learn stochastic neural networks that approximate $\PD$. In this work we build on the GAN framework~\cite{goodfellow2014generative}, which is composed of a generator, $o = G(z)$, mapping a noise input $z\sim \PN(z)$ to an observation, and a discriminator, $D(o)$, mapping an observation to the probability that it was sampled from the real data and not from the generator. The GAN training optimizes a game between the generator and discriminator,
$$
\min_G \max_D V(G, D) = \min_G \max_D \mathbb{E}_{o\sim \PD}\left[ \log D(o) \right] + \mathbb{E}_{z\sim \PN}\left[ \log \left(1 - D(G(z)) \right) \right].
$$

One can view the noise vector $z$ in the GAN as containing some representation of the observation $o$. 
In a general GAN training, however, there is no incentive for this representation to display any structure at all, making it difficult to interpret, or use in a downstream task. The InfoGAN method~\cite{chen2016infogan} aims to mitigate this issue. 

Let $H$ denote the entropy of a random variable $H(x)=\mathbb{E}_{x}\left[ -\log(P(x))\right]$.
The mutual information between the two random variables, $I(x;y) = H(x) - H(x|y)=H(y) - H(y|x)$, measures how much knowing one variable reduces the uncertainty about the other variable.

The idea in InfoGAN is to add to the generator input an additional `state'\footnote{In~\cite{chen2016infogan}, $s$ is referred to as a \emph{code}. Here we term it as a state, to correspond with our subsequent development of structured GAN input from a dynamical system.} component $s\sim P(s)$, and add to the GAN objective a loss that induces maximal mutual information between the generated observation and the state. 
The InfoGAN objective is given by:
\begin{equation}\label{eq:infogan}
\min_{G} \max_D \quad V(G, D) - \lambda I\left(s ; G(z, s)\right),
\end{equation}
where $\lambda>0$ is a weight parameter, and $V(G,D)$ is the GAN loss above. Intuitively, this objective induces
the state to capture the most salient properties of the observation.

Optimizing the objective in \eqref{eq:infogan} directly is difficult without access to the posterior distribution $P(s | o)$, and a variational lower bound was proposed in~\citep{chen2016infogan}. Define an auxiliary distribution $Q(s | o)$ to approximate the posterior $P(s | o)$. Then:
\begin{equation*}\label{eq:InfoGAN_LB}
I\left(s ; G(z, s)\right) \geq 
\mathbb{E}_{\substack{s\sim P(s), o\sim G(z,s)}}\left[\log Q(s | o) \right] + H(s).
\end{equation*}
Using this bound, the InfoGAN objective \eqref{eq:infogan} can be optimized using stochastic gradient descent. Note that the bound is tight when $Q(s|o)$ converges to the true posterior $P(s|o)$.

\subsection{Problem Formulation}
We consider a fully observable and deterministic dynamical system,
% \begin{equation}\label{eq:dyn_sys}
$
o_{t+1} = f(o_t, u_t),    
$
% \end{equation}
where $o_t$ and $u_t$ denote the observation and action at time $t$, respectively.
% \footnote{Typically, formulations for dynamical systems separate the state from the observation, and define a partially observable system, where goals and transitions are defined with respect to the state, and the observation is assumed to be a noisy function of the state~\cite{bertsekas2005dynamic}. For simplicity, here we assume that the observations contain all relevant information about the state, and that the system is deterministic. This allows us to formulate the system with respect to the observation, and ignore the additional technicalities of partial observability. }
The function $f$ is assumed to be unknown. We are provided with data $\data$ in the form of $N$ trajectories of observations $\left\{o_1^{i},u_1^{i}\dots,o_{T_i}^i \right\}_{i\in1,\dots,N},$ generated from $f$, where the actions 
% generating the trajectories are not observed, and 
are generated by an arbitrary exploration policy.\footnote{In this work, we do not concern the problem of how to best generate the exploration data.} 
A typical goal-directed planning problem is the following (e.g., \citep{finn2017deep,agrawal2016learning}):
% \vspace{-0.5em}
\begin{problem}\label{prob:planning}{\textbf{Path Planning}:}
Given $\data$, and two observations $o_{start}, o_{goal}$, generate a sequence of actions that transition the dynamical system $f$ from $o_{start}$ to $o_{goal}$. 
\end{problem}
% \vspace{-0.5em}
For realistic long-horizon planning, however, Problem \ref{prob:planning} can be unnecessarily difficult to solve. As an example, consider a robot planning to navigate through a building. Planning each motor command for the robot in advance seems redundant -- instead, planning a set of way points for the robot and later designing a simple feedback controller to reach them seems much more effective. This concept of \emph{temporal abstraction} has been fundamental in AI planning (e.g.,~\cite{fikes_72,sutton1999between}). To facilitate temporal abstraction in our setting we propose to solve the following, relaxed, problem.
% We assume that $f$ is deterministic.\footnote{In the sense that all stochasticity can be explained by the policy generating the actions.}
%% TK: We might need to use actions to train an inverse model.

We say that two observations $o, o'$ are \emph{$h$-reachable} if there exists a sequence of actions that takes the system from $o$ to $o'$ in $h$ or fewer time steps.
We consider the problem of generating a \emph{walkthrough} -- a sequence of $h$-reachable observations along a feasible path between the start and the goal:
% \vspace{-0.5em}
\begin{problem}\label{prob:reach}{\textbf{Walkthrough Planning}:}
Given $\data$, $h$, and two observations $o_{start}, o_{goal}$, generate a sequence of observations $o_{start},\dots,o_{goal}$ such that every two consecutive observations in the sequence are $h$-reachable. If such a sequence does not exist, return $\emptyset$. 
\end{problem}
% \vspace{-0.5em}
% We also consider a stronger version of this problem which we call the shortest path problem:
% \begin{problem}\label{prob:sp}{\textbf{Shortest Path}:}
% Given $\data$, $h$, and two observations $o_{start}, o_{goal}$, generate the shortest sequence of observations $o_{start},\dots,o_{goal}$ such that every two consecutive observations in the sequence are $h$-reachable. If such a sequence does not exist, return $\emptyset$. 
% \end{problem}
The motivation to solve problem \ref{prob:reach} is that it breaks the long horizon planning problem (from $o_{start}$ to $o_{goal}$) into a sequence of short $h$-horizon planning problems which can be later solved effectively using other methods such as inverse dynamics or model-free RL~\cite{nair2017combining}. 
Note that we are not searching for action sequences, but for a sequence of way point observations. Thus, the actions are not relevant for our problem, and in the sequel we omit them from the discussion.
\section{Causal InfoGAN}\label{s:causal_infogan}
% \vspace{-1em}

A natural approach for solving the walkthrough planning problem in Section \ref{s:problem_formulation} is learning some model of the dynamics $f$ from the data, and searching for a plan within that model. 
This leads to a trade-off. On the one hand, we want to be expressive, and learn all the transitions possible from every $o$ within a horizon $h$. When $o$ is a high dimensional image observation, this typically requires mapping the image to an extensive feature space~\cite{oh2015action,finn2017deep}. On the other hand, however, we want to plan efficiently, which generally requires either low dimensional state spaces or well-structured representations.
We approach this challenge by proposing \textit{Causal InfoGAN} -- an expressive generative model with a structured representation that is compatible with planning algorithms. In this section we present the Causal InfoGAN generative model, and in Section \ref{s:planning} we explain how to use the model for planning.

Let $o$ and $o'$ denote a pair of sequential observations from the dynamical system $f$, and let $\PD(o,o')$ denote their probability, as displayed in the data $\data$. We posit that a generative model that can accurately learn $\PD(o,o')$ has to capture the features that are important for representing the \emph{causality} in the data. By causality here, we mean the next observations $o'$ that are reachable from the current observation $o$. Naturally, such features would be useful later for planning.

We build on the GAN framework~\cite{goodfellow2014generative}. Applied to our setting, a vanilla GAN would be composed of a generator, $o,o' = G(z)$, mapping a noise input $z\sim \PN(z)$ to an observation pair, and a discriminator, $D(o, o')$, mapping an observation pair to the probability that it was sampled from the real data $\data$ and not from the generator. 
% The GAN training (cf. Section \ref{s:problem_formulation}) optimizes the following game between the generator and discriminator,
% $
% \min_G \max_D V(G, D) = \min_G \max_D \mathbb{E}_{o,o'\sim \PD}\left[ \log D(o, o') \right] + \mathbb{E}_{z\sim \PN}\left[ \log \left(1 - D(G(z)) \right) \right].
% $
One can view the noise vector $z$ in such a GAN as a feature vector, containing some representation of the transition to $o'$ from $o$. The problem, however, is that the structure of this representation is not necessarily easy to decode and use for planning. Therefore, we propose to design a generator with a \emph{structured} input that can be later used for planning. In particular, we propose a GAN generator that is driven by states sampled from a parametrized dynamical system.

Let $\model$ denote a dynamical system with state space $\States$, which we term the set of \textit{abstract-states}, and a parametrized, stochastic transition function $T_{\model}(s'|s)$:
% \begin{equation}
$
    s' \sim T_{\model}(s'|s),
$
% \end{equation}
where $s,s'\in\States$ are a pair of consecutive abstract states. We denote by $P_{\model}(s)$ the prior probability of an abstract state $s$.
% \footnote{By abuse of notation, we denote all the parameters in our model by $\theta$. However, we emphasize that each component in the model may depend on different elements of $\theta$.} 
We emphasize that the abstract state space $\States$ can be different from the space of real observations $o$. For reasons that will become clear later on, we term $\model$ as the \emph{latent planning system}.

We propose to structure the generator as taking in a pair of consecutive abstract states $s,s'$ in addition to the noise vector $z$. The GAN objective in this case is therefore (cf. Section \ref{s:problem_formulation}):
\begin{equation}\label{eq:GAN_objective_model}
    V(G, D) = \mathbb{E}_{o,o'\sim \PD}\left[ \log D(o, o') \right] + \mathbb{E}_{z\sim \PN,s\sim P_{\model}(s),s'\sim T_{\model}(s)}\left[ \log \left(1 - D(G(z,s,s')) \right) \right].
\end{equation}

The idea is that $s$ and $s'$ would represent the abstract features that are important for understanding the causality in the data, while $z$ would model variations that are less informative, such as pixel level details. To induce learning such representations, we follow the InfoGAN method~\cite{chen2016infogan}, and add to the GAN objective a loss that induces a maximal mutual information between the generated pair of observations and the abstract states.

% Let $H$ denote the entropy of a random variable $H(x)=\mathbb{E}_{x}\left[ -\log(P(x))\right]$.
% The mutual information between the two random variables, $I(x;y) = H(x) - H(x|y)=H(y) - H(y|x)$, measures how much knowing one variable reduces the uncertainty about the other variable.
% the reduction in uncertainty of one variable that is obtained by knowing the other variable.
% $I(s,s';o,o')=H(s,s') - H(s,s'|o,o')=H(o,o') - H(o,o'|s,s')$
% denotes the mutual information between the pair of states $s,s'$ and generated observations $o,o'$. 
% In our case, we would like the transitions in $\model$ to convey information about the generated observation transitions, and 
We propose the Causal InfoGAN objective:
\begin{equation}\label{eq:causal_infogan}
\begin{split}
    % \min_{\model,G} \max_D &\quad V(G, D) - \lambda \left(I\left(s ; o\right) + I\left(\left. s'; o'\right|s, o\right)\right), \\
    \min_{\model,G} \max_D &\quad V(G, D) - \lambda I\left(s, s' ; o, o'\right), \\
    % \min_{\model,G} \max_D &\quad \mathbb{E}_{o,o'\sim \PD}\left[ \log D(o, o') \right] + \mathbb{E}_{z,s,s'}\left[ \log \left(1 - D(G(z,s,s')) \right) \right] - \lambda I\left(s, s' ; o, o'\right), \\
    \textrm{s.t. } &\quad o,o' \sim G(z, s, s') \quad 
                   \quad s\sim P_{\model}, \quad 
                   \quad s'\sim T_{\model}(s), 
\end{split}
\end{equation}
where $\lambda>0$ is a weight parameter, and $V(G,D)$ is given in \eqref{eq:GAN_objective_model}.
%% TK: *** I think it is important to make it clear that the mutual information between (s and o) and (s' and o') are disjoint. It doesn't make sense to do them jointly because we want s to represent the state of o in order to do planning. 
%% TK: I saw that we have disentangled classifier later on, but I still think it's  good to explain to the reader that s corresponds to o and s' corresponds to o'.
%% AT: good point. I modified the text to reflect this. 
Intuitively, this objective induces
the abstract model to capture the most salient possible changes that can be effected on the observation.

Optimizing the objective in \eqref{eq:causal_infogan} directly is difficult, since we do not have access to the posterior distribution, $P(s, s' | o, o')$, when using an expressive generator function. Following InfoGAN~\citep{chen2016infogan}, we optimize a variational lower bound of \eqref{eq:causal_infogan}. Define an auxiliary distribution $Q(s, s' | o, o')$ to approximate the posterior $P(s, s' | o, o')$. We have, following a similar derivation to \citep{chen2016infogan}:
\begin{equation}\label{eq:I_LB}
I\left((s, s') ; G(z, s, s')\right) \geq 
\mathbb{E}_{\substack{s\sim P_{\model},  s'\sim T_{\model}(s), \\o,o'\sim G(z,s,s')}}\left[\log Q(s, s' | o, o') \right] + H(s,s') \doteq I_{VLB}(G,Q).
\end{equation}
% where $H(x)=\mathbb{E}_{x}\left[ -\log(P(x))\right]$ denotes the entropy.
In this formulation, $Q$ can be seen as a classifier, mapping pairs of observations to pairs of abstract states. 

% \TODO{The sensitivity is confusing to me. Can we say that because the mutual information does not encourage $s$ to capture the abstraction of $o$ so we need an additional assumption?}
We now note a subtle point. The mutual information in \eqref{eq:causal_infogan} is not sensitive to the order of the code words of the random variables $s$ and $s'$.\footnote{This is a general property of the entropy of a random variable, which only depends on the probability distribution and not on the variable values. In our case, for example, one can apply a permutation to the transition operator $T_{\model}$, and an inverse of that permutation to the generator's $s'$ input. Such a permutation would change the meaning of $s'$, without changing the mutual information term nor the distribution of generated observations.} 
This points to a potential caveat in the optimization objective \eqref{eq:causal_infogan}: we would like the random variable for the next abstract state $s'$ to have the \emph{same meaning} as the random variable for the abstract state $s$. That would allow us to roll-out a sequence of changes to the abstract state, by applying the transition operator $T_{\model}$ sequentially, and effectively plan in the abstract model $\model$. 
% However, the loss function \eqref{eq:causal_infogan} does not automatically induce $s'$ to have the same meaning as $s$, since 
We solve this problem by proposing the  \emph{disentangled posterior approximation}, $Q(s, s' | o, o') = Q_1(s|o)Q_2(s'|o')$, and choose $Q_1=Q_2\doteq Q$. This effectively induces a generator for which $P(s|o)=P(s'|o')$.\footnote{Note that in a system where the state is fully observable, the posterior is disentangled by definition, therefore in such cases the bound is tight.}
% , as we explain next.
% To promote $s$ and $s'$ to have the same meaning, 
% \TODO{We want to say that the bound is tight. When is this the case in real posterior? At least true in fully observable case.}. 

% The bound in \eqref{eq:I_LB} is tight, and 
We use the lower bound \eqref{eq:I_LB} in \eqref{eq:causal_infogan} to obtain the following loss function:
\begin{equation}\label{eq:CI_loss_vlb}
    \min_{G,Q,\model} \max_D \quad V(G, D) - \lambda I_{\text{VLB}}(G,Q),
\end{equation}
% \begin{equation}\label{eq:CI_loss_vlb}
%     \min_{G,Q,\model} \max_D \quad V(G, D) - \lambda_{\text{VLB}} I_{\text{VLB}}(G,Q),
% \end{equation}
% where $\lambda_{\text{VLB}}$ is a constant. The loss in \eqref{eq:CI_loss_vlb} can be optimized effectively using stochastic gradient descent.
where $\lambda>0$ is a constant. The loss in \eqref{eq:CI_loss_vlb} can be optimized effectively using stochastic gradient descent, and we provide a detailed algorithm in Appendix \ref{sec:algorithm}.

%\TODO{Change this loss to regularization term: we just induce small transition variance in continuous case and encourage clusters to stay the same in discrete case. Again the reason that SC loss has the same effect is that Q tends to map to a few modes given real observations.}
%\TODO{Add a section improvement to the loss function after this section. Show this in ablation study.}

\begin{figure}[h!]
    \centering
    % \hfill
    \begin{subfigure}[b]{0.71\textwidth}
        \centering
        \includegraphics[width=\textwidth]{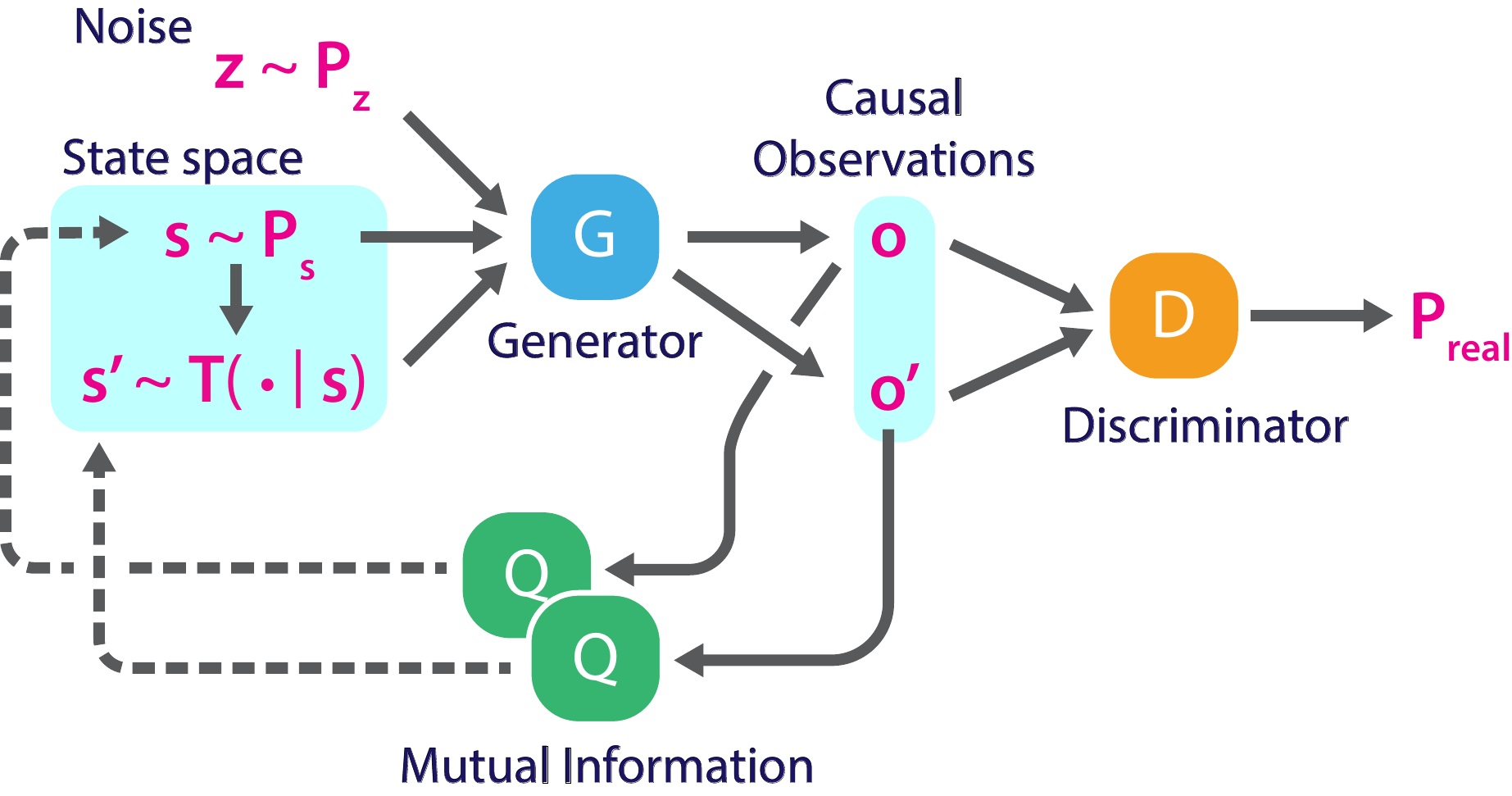}
        \caption{Causal InfoGAN model}
    \end{subfigure}
    % \hfill
    \begin{subfigure}[b]{0.28\textwidth}
        \centering
        \includegraphics[width=\textwidth]{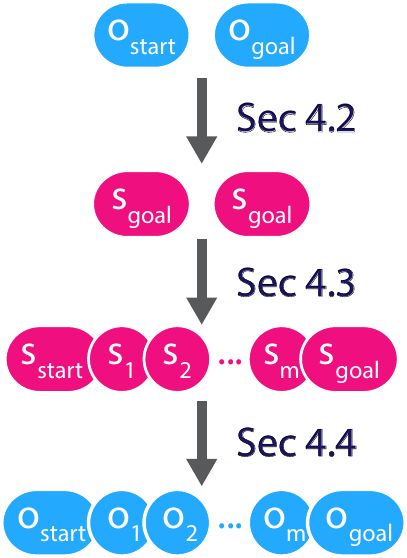}
        \caption{Planning paradigm}\label{fig:planning_paradigm}
    \end{subfigure}
    \caption{The Causal InfoGAN framework. \textbf{(a) Generative model (cf. Section \ref{s:causal_infogan})}. First, an abstract state $s$ is sampled from a prior $P_\model(s)$. Given $s$, the next state $s'$ is sampled using the transition model $T_\model(s'|s)$. The states $s, s'$ are fed, together with a random noise sample $z$, into the generator which outputs $o, o'$. The discriminator $D$ maps an observation pair to the probability of the pair being real. Finally, the approximate posterior $Q$ maps from each observation to the distribution of the state it associates with. The causal InfoGAN loss function in Equation \eqref{eq:CI_loss_vlb} encourages $Q$ to predict each state accurately from each observation. 
    % Note that the two posterior functions are identical. They compute the likelihood of the state label for $o$ and $o'$ independently. 
    \textbf{(b) Planning paradigm (cf. Section \ref{s:planning})}. Given start and goal observations, we first map them to abstract states, and then we apply planning algorithms using the model $\model$ to search for a path from $s_{start}$ to $s_{goal}$. Finally, from the plan in abstract states, we generate back a sequence of observations.
    }
    \label{fig:schematic}
    % \vspace{-1em}
\end{figure}

\section{Planning with Causal InfoGAN models}\label{s:planning}

In the previous section, we proposed a general framework for learning a deep generative model of data from a dynamical system, with a structured latent space. In this section, we discuss how to use the Causal InfoGAN model for planning goal directed trajectories. We first present our general methodology, and then propose several model configurations for which \eqref{eq:CI_loss_vlb} can be optimized efficiently using backpropagation and the reparametrization trick~\cite{chen2016infogan}, and the latent planning system is compatible with efficient planning algorithms. We then describe how to combine these ideas for solving the walkthrough planning problem in various domains.
% later explain how Causal InfoGAN can be used to solve the planning problem \ref{prob:reach}.

% Our model abstracts away the raw observations by learning high-level \textit{abstract-states} and transitions among abstract-states from the raw data $\data$. 

% 
%
\subsection{General Planning Paradigm}
Our general paradigm for goal directed planning is described in Figure ~\ref{fig:planning_paradigm}. We start by training a Causal InfoGAN model from the data, as described in the previous section. Then, we perform the following 3 steps, which are detailed in the rest of this section:
\begin{enumerate}
    \item Given a pair of observations $o_{start}, o_{goal}$, we first encode them into a pair of corresponding states $s_{start}, s_{goal}$. This is described in Section \ref{ss:encode_state_to_obs}.
    \item Then, using the transition probabilities in the planning model $\model$, we plan a trajectory $s_{start}, s_1,\dots, s_m, s_{goal}$ -- a feasible sequence of states from $s_{start}$ to $s_{goal}$. This is described in Section \ref{ss:latent_systems}.
    \item Finally, we decode the state trajectory into a corresponding trajectory of observations $o_{start}, o_1,\dots, o_m, o_{goal}$. This is described in Section \ref{ss:decoding_trajectory}.
\end{enumerate}  

% \paragraph{Consistent abstract states:} 
In order for the planned trajectory to be consistent with Problem \ref{prob:reach}, any two consecutive observations that correspond to consecutive abstract-states, i.e., states that can be reached in a single transition, have to be $h$-reachable. To train such $h$-reachable abstract states, we simply train the Causal InfoGAN model with pairs of observations $o, o'$ from $\data$ that are separated by at most $h$ time steps. 
% \TODO{not sure this paragraph is best here. Thoughts? I think this is good to put here. We can further emphasize here the generality of our method for temporal abstraction.}

The specific method for each step in the planning paradigm can depend on the problem at hand. For example, some systems are naturally described by discrete abstract states, while others are better described by continuous states. In the following, we describe several models and methods that worked well for us, under the general planning paradigm described above. This list is by no means exhaustive. On the contrary, we believe that the Causal InfoGAN framework provides a basis for further investigation of deep generative models that are compatible with planning.

\subsection{Encoding an Observation to a State}\label{ss:encode_state_to_obs}

For mapping an observation to a state, we can simply use the disentangled posterior $Q(s|o)$. We found this approach to work well in low-dimensional observation spaces. However, for high-dimensional image observations we found that the learned $Q(s|o)$ was accurate in classifying generated observations (by the generator), but inaccurate for classifying real observations. This is explained by the fact that in Causal InfoGAN, $Q$ is only trained on generated observations, and can therefore overfit to generated images.

In high-dimensional domains, we therefore opted for a different approach. Following \cite{wang2017safer}, we performed a search over the latent space to find the best latent state mapping $s^*(o)$:
\begin{equation*}
    s^*(o) = \argmin_{s} \min_{s',z} \| o - G(s,s',z) \|^2.
\end{equation*}

Another approach, which could scale to complex image observations, is to add to the GAN training an explicit encoding network~\cite{donahue2016adversarial,zhu2017toward}. In our experiments, the simple search approach worked well and we did not require additional modifications to the GAN training.

\subsection{Latent Planning Systems}\label{ss:latent_systems}

We now present several latent planning systems that are compatible with efficient planning algorithms. Table \ref{table:state_reps} summarizes the different models.

\begin{table}
\centering
\begin{tabular}{ |c|c|c|c|c| } 
 \hline
 Type & Values $s$ & Prior $P_{\model}(s)$ & Transition $T_{\model}(s'|s)$ & Planning algorithms \\ 
 \hline
 Discrete -- one-hot & $[N]$ & $\mathcal{U}\{1,\dots,N\}$ & $s'\sim Softmax(s^{\top} \theta)$ & Dijkstra \\ 
 Discrete -- binary & $\{0,1\}^N$ & $\mathcal{U}\{0,1\}^N$ & See eq. \ref{eq:binary_trans} & Dijkstra\\ 
 Continuous & $\mathbb{R}^N$ & $\mathcal{U}(-1,1)^N$ & $s'\sim\mathcal{N}(s, \Sigma_\theta(s))$ & Linear interpolation\\ 
 \hline
\end{tabular}
\vspace{1em}
\caption{Different models for the latent planning system. In all cases, $N$ is the state dimension. The parameters $\theta$ of the transition $T_\model$ have different forms depending on the state types. In the one-hot case, $\theta$ is a matrix in $\mathbb{R}^{N\times N}$. In the binary case, $\theta$ denotes parameters in a stochastic neural network; see Eq. \eqref{eq:binary_trans}. In the continuous case $\theta$ represents the parameters of a neural network that controls the variance of the transition.}\label{table:state_reps}
\end{table}

\subsubsection{Discrete Abstract States -- One-Hot Representation}\label{s:state_agg}

% Consider a finite Markov decision process (MDP,~\cite{sutton1998reinforcement}) model for $\model$ with $N$ states, 
We start from a simple abstract state representation, in which each $s\in S$ is represented as a $N-$dimensional one-hot vector. We denote by $\theta\in \mathbb{R}^{N\times N}$ the model parameters, and compute transition probabilities as:
% \begin{equation*}
$
    T_{\model}(s'|s) = Softmax(s^{\top}\theta) .
$
% \end{equation*}
% The actions $a$ in this case are implicitly represented by the noise in the Gumbel softmax distribution. 
Optimizing the parameters $\theta$ with respect to the expectation in the loss \eqref{eq:I_LB} is done using the Gumbel-softmax reparametrization trick~\cite{jang2016categorical}.

% Solving for the shortest path in this model can easily be done using Dijkstra's algorithm, on a directed graph where states $s, s'$ are defined to be connected if $P(s'|s) > \epsilon$, where $\epsilon$ is a hyperparameter.

\subsubsection{Discrete Abstract States -- Binary Representation}\label{s:binary_state}
We now present a more expressive abstract state representation using binary states. Binary state representations are common in AI planning, where each binary element is known as a \emph{predicate}, and corresponds to a particular property of an object being true or false~\cite{russel2010AI}. The Causal InfoGAN framework allows us to learn the predicates \emph{directly from data}. 

We propose a parametric transition model that is suitable for binary representations. Let $s\in \{0,1\}^{N}$ be an $N-$dimensional binary vector, drawn from $P_{\model}(s)$. We generate the next state $s'$ by first drawing a random \emph{action vector} $a\in \{0,1\}^{M}$ with some probability $P{_\model}(a)$. The purpose of this random action is to generate stochasticity in the state transition.
Let $MLP(s,a)$ denote a multi-layered perceptron with parameters $\theta$ mapping the state $s$ and action $a$ to $\mathbb{R}^N$. The probability of the next state $s'$ is finally given by:
\begin{equation}\label{eq:binary_trans}
% \begin{split}
    T_{\model}(s'=v|s)  = \mathbb{E}_{a} \left[\prod_i T_{\model}(s'_i=v_i|s,a) \right], \textrm{where }
    T_{\model}(s'_i=1|s,a)  = Sigmoid(MLP(s,a)_i).
% \end{split}
\end{equation}
Thus, for a given action, each element in $MLP(s,a)$ can be interpreted as the logit in a binary distribution for generating the corresponding element in $s'$, and for calculating the state transition probability we marginalize over the action. Note that the binary distributions for the different elements in $s'$ are independent \emph{given $s$ and $a$}. Thus, for a particular $s$, complex distributions for $s'$ may be expressed through the MLP dependence on $a$. We further emphasize that there is not necessarily any correspondence between the action vector $a$ and the real actions that generated the observation pairs in the data. The action $a$ is simply a means to induce stochasticity to the state transition network.
Optimizing the parameters $\theta$ with respect to the expectation in the loss \eqref{eq:I_LB} is done using the Gumbel-softmax trick~\cite{jang2016categorical} for each element in the MLP output. In this work, we chose $P_{\model}(s)$ and $P_{\model}(a)$ to be fixed distributions, where each binary element was independent, with a Bernoulli$(0.5)$ distribution. In this case, the marginalization can be calculated in closed form. It is also possible to extend this model to a parametric distribution for $P_{\model}(s)$ and $P_{\model}(a)$, and marginalize using sampling.
% The actions in this case, as in section \ref{s:state_agg}, are implicitly represented by the noise in the Gumbel softmax distribution.

% \AT{Should we discuss other possible models in appendix? E.g., schema networks, linear quadratic regulators. Would strengthen the generality of the framework. If we have time/space we can do some experiments with soft schema networks in the paper.}

Both the one-hot and binary representations defined above can be seen as learning a finite Markov decision process (MDP, \cite{bertsekas2005dynamic}) model of the data. In the one-hot case, actions in the MDP are implicit in the Gumbel softmax noise, while in the binary case, they are explicit. This is, in fact, a form of state aggregation~\cite{bertsekas2005dynamic}, and we can think of $Q(s|o)$ as a function that assigns a soft clustering to the observations. In contrast to standard clustering approaches in the literature \cite{simester2006dynamic,mahadevan2007proto,lakshminarayanan2016option,baram2016spatio}, our method does not require a metric function on the observation space, nor a value function, which depends on a particular task through the reward function. We illustrate these advantages in our experiments.

For planning with discrete models, we interpret the stochastic transition model $T_{\model}$ as providing the possible state transitions, i.e., for every $s'$ such that $T_{\model}(s'|s) > \epsilon$ there exists a possible transition from $s$ to $s'$. For planning, we require abstract state representations that are compatible with efficient AI planning algorithms. The one-hot and binary representations above can be directly plugged in to graph-planning algorithms such as Dijkstra's shortest-path algorithm~\cite{russel2010AI}. 

\subsubsection{Continuous Abstract States}\label{s:continuous_state}
For some domains, such as the rope manipulation in our experiments, a continuous abstract state is more suitable. We consider a model where an $s\in S$ is represented as a $N-$dimensional continuous vector. Planning in high-dimensional continuous domains, however, is hard in general. 

Here, we propose a simple and effective solution: we will learn a latent planning system such that \emph{linear interpolation between states makes for feasible plans}. To bring about such a model, we consider transition probabilities $T_{\model}(s'|s)$ given as Gaussian perturbations of the state:
% \begin{equation*}
$
    s' = s + \delta,
$
% \end{equation*}
% The actions $a$ in this case are implicitly represented by the noise in the Gumbel softmax distribution. 
where $\delta \sim \mathcal{N}(0,\Sigma_\theta(s))$ and $\Sigma_\theta$ is a diagonal covariance matrix, and is represented by a MLP with parameters $\theta$. The key idea here is that, if only small local transitions are possible in the system, then a linear interpolation between two states $s_{start}, s_{goal}$ has a high probability, and therefore represents a feasible trajectory in the observation space. To encourage such small transitions, we add an L2 norm of the convariance matrix to the full loss \eqref{eq:CI_loss_vlb}.

\begin{equation}\label{eq:lcont}
    L_{cont}(\model)=\mathbb{E}_{s\sim P_{\model}}\norm{ \Sigma_\theta(s) }_2
\end{equation}

The prior probability $P_{\model}$ for each element of $s$ is uniform in $[-1,1]$.
Optimizing the parameters $\theta$ with respect to the expectation in the loss \eqref{eq:I_LB} is done using the reparametrization trick~\cite{kingma2013auto}. 
% Note that this model induces a latent representation in which local changes to the abstract state correspond to feasible changes to the observation. Thus, we can plan in this model by simply interpolating between abstract states.

\subsection{Decoding a State Trajectory to an Observation Walkthrough Trajectory}\label{ss:decoding_trajectory}

We now discuss how to generate a feasible sequence of observations from a state trajectory in the latent planning system. Here, as before, we separate the discussion for systems with low-dimensional observations and systems with high-dimensional observations, as we found that different practices work best for each.

For low-dimensional observations, we structure the GAN generator $G$ to have an observation-conditional form:
\begin{equation}
    % \begin{split}
        o = G_1(z, s, s'), \quad
        o' = G_2(z, o, s, s').
    % \end{split}
\end{equation}
Using this generator form, we can sequentially generate observations from a state sequence $s_1,\dots,s_T$. We first use $G_1$ to generate $o_1$ from $s_1,s_2$, and then, for each $2\leq t < t$, use $G_2$ to generate $o_{t+1}$ from $s_t, s_{t+1}$, and $o_t$. 

% To further improve the planning result and generate more realistic trajectories, we have found that we can use the GAN discriminator $D$ to provide a confidence score for the resulting trajectory, and prune generated trajectories that have low confidence. Therefore, we generate $K$ random trajectories using the generator as described above, with different random noise $z$, and choose the one with the best average score. % as $\argmax_{i\in 1,\dots,K}\sum_t D(o_t^i,o_{t+1}^i)$.

For high-dimensional image observations, the sequential generator does not work well, since small errors in the image generation tend to get accumulated when fed back into the generator. We therefore follow a different approach.
To generate the $i$'th observation in the trajectory $o_i$, we use the generator with the input $s_i, s_{i+1}$, and a noise $z$ that is fixed throughout the whole trajectory. The generator actually outputs a pair of sequential images, but we discard the second image in the pair.

% As in the low-dimensional case, to further improve the planning result we generate $K$ random trajectories with different random noise $z$, and select the best trajectory by using a discriminator $D$ to provide a confidence score for each trajectory. 
% Here, we found that the GAN discriminator is not reliable enough, and trained an auxiliary discriminator for novelty detection, as described in the experiments section.\TODO{Thanard - make sure that what I wrote is actually what we did.}
To further improve the planning result we generate $K$ random trajectories with different random noise $z$, and select the best trajectory by using a discriminator $D$ to provide a confidence score for each trajectory. In the low-dimensional case, we use the GAN discriminator. In the high-dimensional case, however, we find that the discriminator tends to overfit to the generator. Therefore, we trained an auxiliary discriminator for novelty detection, as described in the Experiment Section \ref{sec:rope-manipulation}.

\section{Related Work}

Combining deep generative models with structured dynamical systems has been  explored in the context of variational autoencoders (VAEs), where the latent space was continuous~\cite{chung2015recurrent,johnson2016svae}. Watter et al.~\cite{watter2015embed} have suggested to use such models for planning, by learning latent linear dynamics, and using a linear quadratic Gaussian control algorithm for planning. 
Disentangled video prediction~\cite{denton2017unsupervised} separates object content and position, but has not been used for planning. 
Very recently, Corneil et al.~\cite{corneil2018efficient} suggested Variational
State Tabulation (VaST) -- a VAE-based approach for learning latent dynamics over binary state representations, and planning in the latent space using prioritized sweeping to speed up RL. 
Causal InfoGAN shares several similarities with VaST, such as using Gumbel-Softmax to backprop through transitions of discrete binary states, and leveraging the structure of the binary states for planning. However, VaST is formulated to require the agent actions, and is thus limited to single time step predictions. More generally, our work is developed under the GAN formulation, which, to date, has several benefits over VAEs such as superior quality of image generation~\cite{karras2017progressive}. Causal InfoGAN can also be used with continuous abstract states.

The semiparametric topological memory (SPTM)~\cite{savinov2018semi} is another  recent approach for solving problems such as Problem \ref{prob:reach}, by planning in a graph where every observation in the data is a node, and connectivity is decided using a learned similarity metric between pairs of observations. SPTM has shown impressive results on image-based navigation. However, Causal InfoGAN's \emph{parametric approach} of learning a compact, model for planning has the potential to scale up to more complex problems, in which the increasing amount of data required would make the nonparametric SPTM approach difficult to apply. 

%%PA.5.18: add a short paragraph on video prediction work that tries to learn disentangled representations?
%% e.g.: (and some references here-in)
% @incollection{NIPS2017_7028,
%title = {Unsupervised Learning of Disentangled Representations from Video},
%author = {Denton, Emily L and Birodkar, vighnesh},
%booktitle = {Advances in Neural Information Processing Systems 30},
%editor = {I. Guyon and U. V. Luxburg and S. Bengio and H. Wallach and R. Fergus and S. %Vishwanathan and R. Garnett},
%pages = {4414--4423},
%year = {2017},
%publisher = {Curran Associates, Inc.},
%url = {http://papers.nips.cc/paper/7028-unsupervised-learning-of-disentangled-representations-from-video.pdf}
%}

% \paragraph{VAE approach}\cite{corneil2018efficient}
% without planning: SVAE \cite{johnson2016svae} and 

% \paragraph{Semi-parametric topological memory for navigation} \cite{savinov2018semi}

Learning state aggregation and state representation has a long history in RL. Methods such as in \citep{mannor2004dynamic,simester2006dynamic} exploit the value function for measuring state similarity, and are therefore limited to the task defined by the reward. Methods for general state aggregation have also been proposed, based on spectral clustering \citep{mahadevan2007proto,lakshminarayanan2016option,machado2017laplacian,liu2017eigenoption}, and variants of K-means \citep{baram2016spatio}. All these approaches rely in some form on the Euclidean distance as a metric between observation features. As we show in our experiments, the Euclidean distance can be unsuitable even on low-dimensional continuous domains.

Recent work in deep RL explored learning goal-conditioned value functions and policies~\cite{andrychowicz2017hindsight,Pong2018TDM}, and policies with an explicit planning computation~\cite{tamar2016value,oh2017value,srinivas2018universal}. These approaches require a reward signal for learning (or supervision from an expert~\cite{srinivas2018universal}). In our work, we do not require a reward signal, and learn a general model of the dynamical system, which is used for goal-directed planning.
% our experiments, we show why such approaches are not suitable when the \citep{simester2006dynamic} cluster states based on the value function. In our case, we do not have rewards so that method does not apply.

Our work is also related to learning models of intuitive physics. Previous work explored feedforward neural networks for predicting outcomes of physical experiments~\cite{lerer2016learning}, neural networks for modelling relations between objects \cite{watters2017visual,santoro2017simple}, and prediction based on physics simulators \cite{battaglia2013simulation,wu2017learning}. To the best of our knowledge, these approaches cannot be used for planning, which is the focus in this paper. However, related ideas would likely be required for scaling our method to more complex domains, such as manipulating several objects.

Using the mutual information as a signal that drives prediction in dynamical systems has also been explored under a different formulation in the information bottleneck line of work~\cite{tishby2000information,amir2015past}. 

In the planning literature, most studies relied on manually designed state representations. In a recent work, Konidaris et al.~\cite{konidaris2018skills} automatically extracted state representations from raw observations, but relied on a prespecified set of skills for the task. In our work, we automatically extract state representations by learning salient features that describe the causal structure of the data.
% \TODO{Something from planning literature? e.g.~\cite{jimenez2012review}?}

\section{Experiments}

In our experiments, we aim to (1) visualize the abstract states and planning in Causal InfoGAN; (2) compare Causal-InfoGAN with recent state-aggregation methods in the literature; (3) show that Causal InfoGAN can produce realistic visual plans in a complex dynamical system; and (4) show that Causal InfoGAN significantly outperforms baseline methods.

% (3) show that Causal InfoGAN can be used to learn and plan in a complex dynamical system with visual observations.

% \begin{itemize}
%     \item Visualize the abstract states and planning in Causal InfoGAN
%     \item Compare Causal-InfoGAN with recent state-aggregation methods in the literature.
%     % \item How does Causal InfoGAN compare with methods for state aggregation?
%     % \item Can planning be done reliably in latent space 
%     % \item Can Causal InfoGAN accurately generate observation trajectories from the plan?
%     % \item Can Causal InfoGAN work on a challenging image-based task of rope manipulation?
% \end{itemize}
We begin our investigation with a set of toy tasks, specifically designed to demonstrate the benefits of Causal InfoGAN, where we can also perform an extensive quantitative evaluation. We later present experiments on a real dataset of robotic rope manipulation. Technical details for reproducing the experiments are provided in the supplementary material. Code will be made available online at \url{http://github.com/thanard/causal-infogan}.

\subsection{Illustrative Experiments}
In this section we evaluate Causal InfoGAN on a set of 2D navigation problems. These toy problems abstract away the challenges of learning visual features, and allow us to make an informative comparison on the task of learning causal structure in data, and using it for planning. For details of the training data see Appendix \ref{sec:experiment_details}.

\begin{figure}[h]
    \centering
    \hfill
    \begin{subfigure}[b]{0.25\linewidth}
    \includegraphics[width=1.0\textwidth]{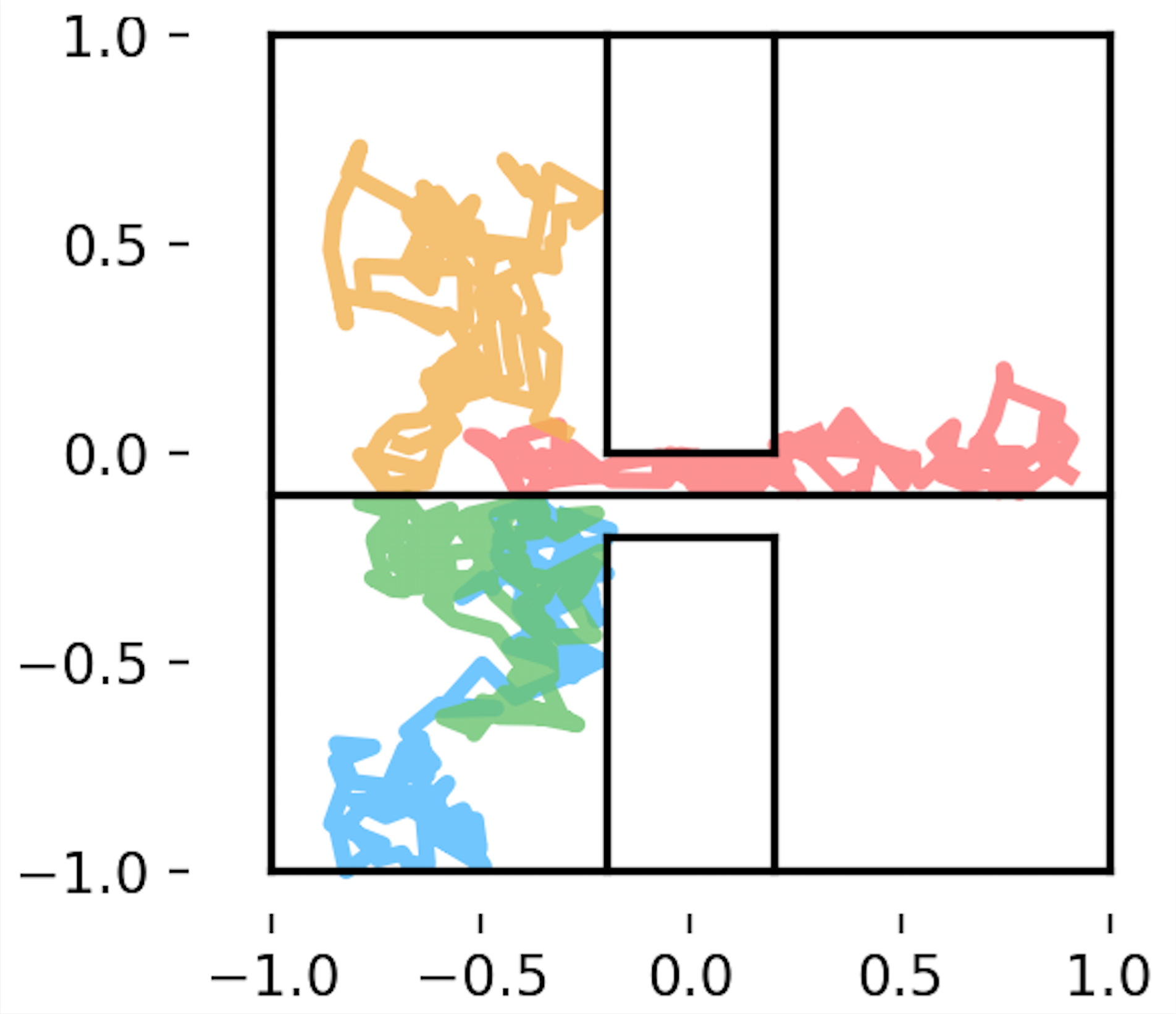}
    \caption{}
    \label{fig:tsp_results}
  \end{subfigure}\hfill
  \begin{subfigure}[b]{0.25\linewidth}
    \includegraphics[width=0.9\textwidth]{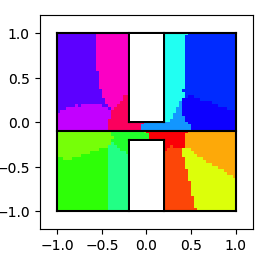}
    \caption{}
  \end{subfigure}\hfill
  \begin{subfigure}[b]{0.25\linewidth}
    \includegraphics[width=0.9\textwidth]{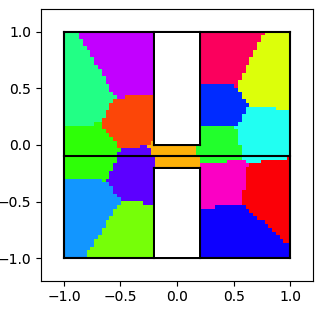}
    \caption{}
  \end{subfigure}
  \begin{subfigure}[b]{0.24\linewidth}
    \includegraphics[width=0.9\textwidth]{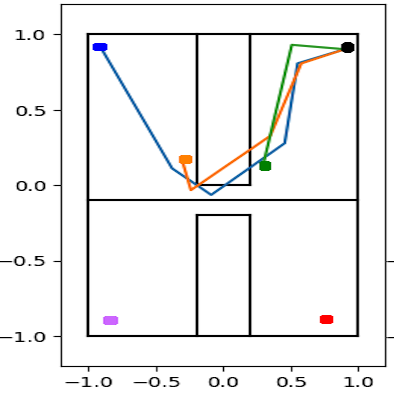}
    \caption{}
  \end{subfigure}
    \caption{2D particle results on tunnel domain. (a) The domain - top/bottom rooms are not connected. Left/right quadrants are connected through a narrow tunnel. An example of several random walk trajectories are shown. (b) Clustering found by Causal InfoGAN. (c) Clustering found by K-means. (d) Example walkthrough trajectories generated by Causal InfoGAN, from a point at the top right to five other locations on the map, marked by colored circles. For trajectories that were not found only the target is shown. Note that Causal InfoGAN learned clusters that correspond to the possible dynamics of the particle in the task, and was therefore able to generate reasonable planning trajectories.}%\TODO{Add key domain figures.}}
    % \AT{This is just a placeholder for real results now.}(a) Tunnel domain, and a sample of data trajectories. (b) Causal InfoGAN abstract states. (c) K-means clustered states. 
    % \GY{Want to add Key-domain. Show planning result next to the cluster.}}
    \label{fig:toy_task}
\end{figure}

\begin{figure}[h]
    \centering
    %\hfill
    \begin{subfigure}[b]{0.205\linewidth}
    \raisebox{2pt}{
    \includegraphics[width=\textwidth]{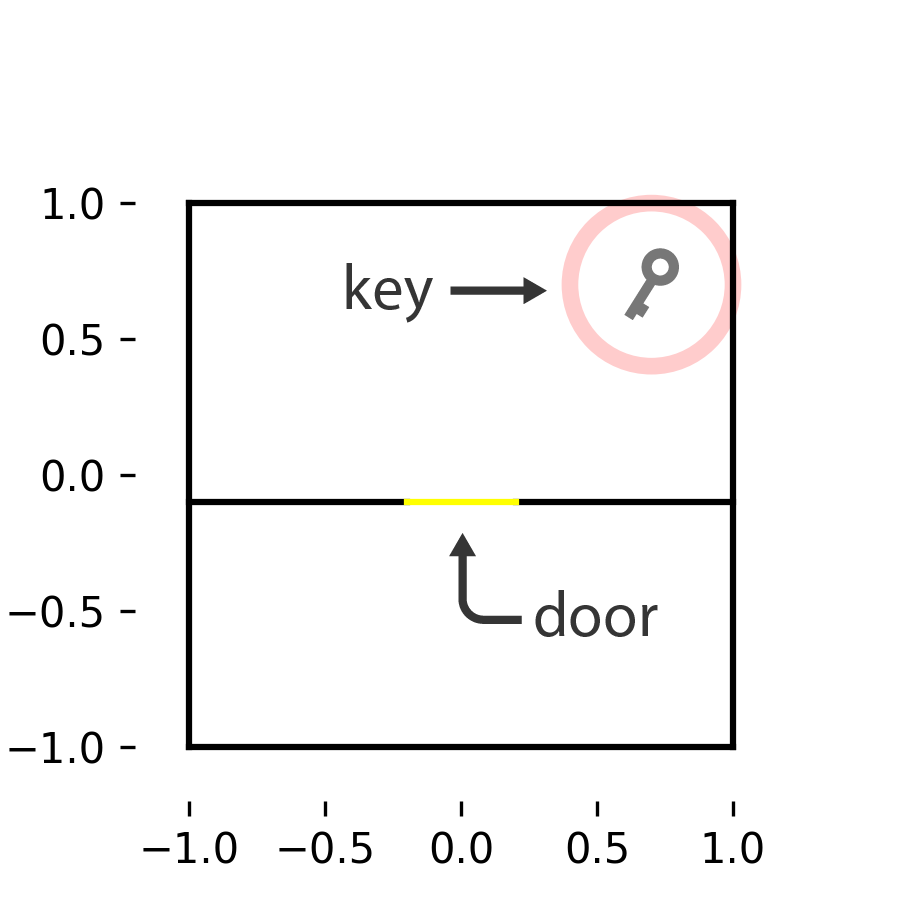}
    }
    \caption{}
    \label{fig:tsp_results}
  \end{subfigure}
  \begin{subfigure}[b]{0.24\linewidth}
    \includegraphics[width=\textwidth]{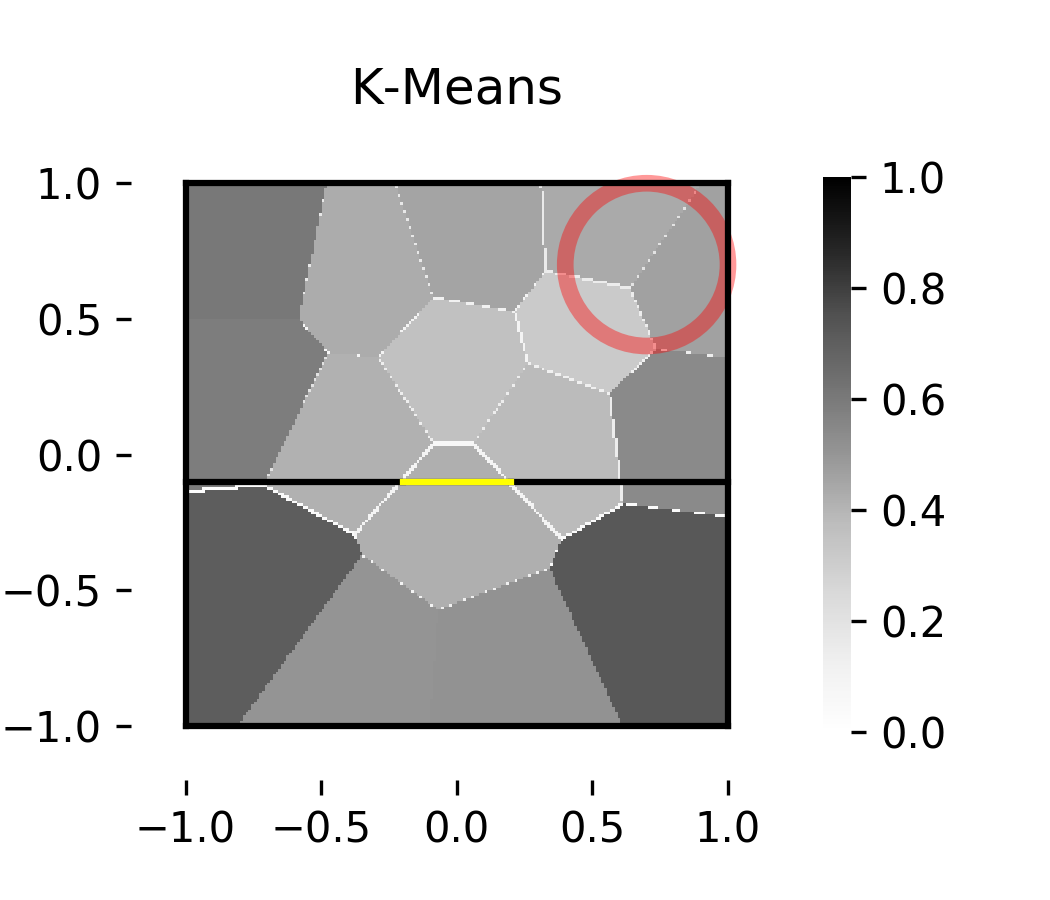}
    \caption{}
  \end{subfigure}
  \begin{subfigure}[b]{0.24\linewidth}
    \includegraphics[width=\textwidth]{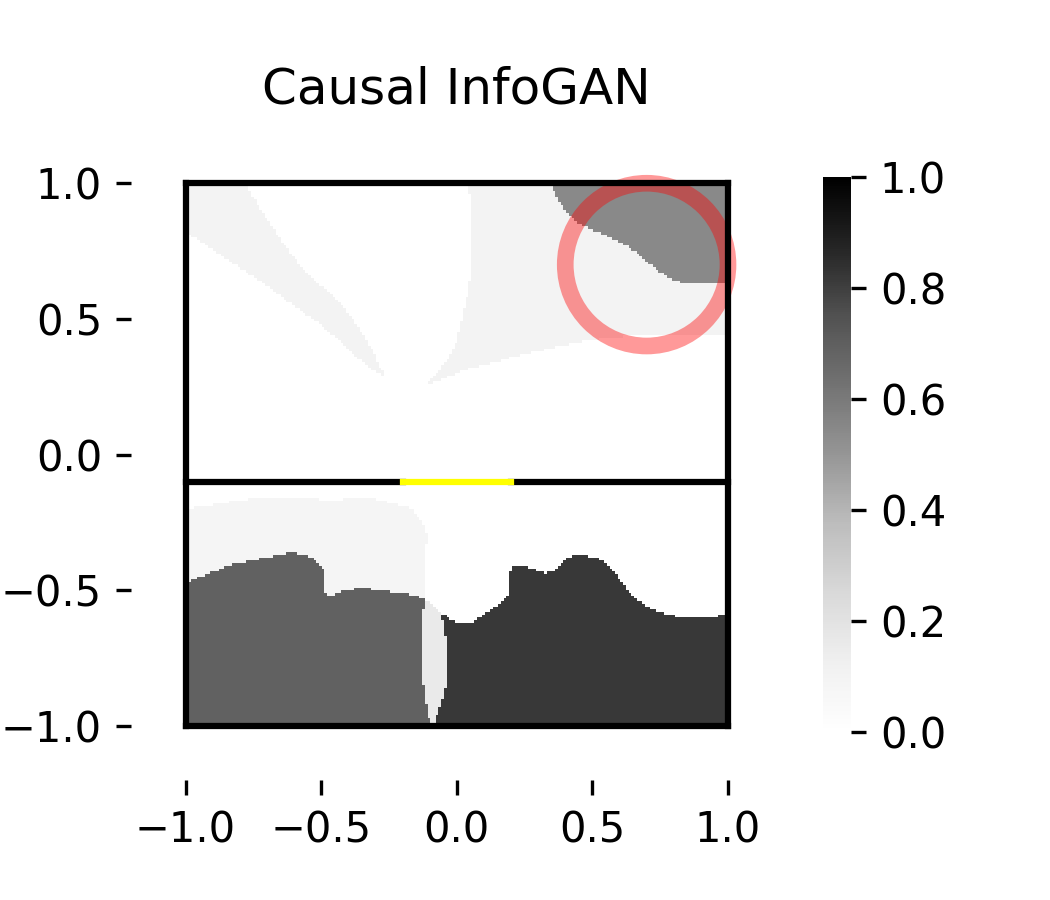}
    \caption{}
  \end{subfigure}
  \begin{subfigure}[b]{0.21\linewidth}
    \raisebox{1pt}{
    \includegraphics[width=\textwidth]{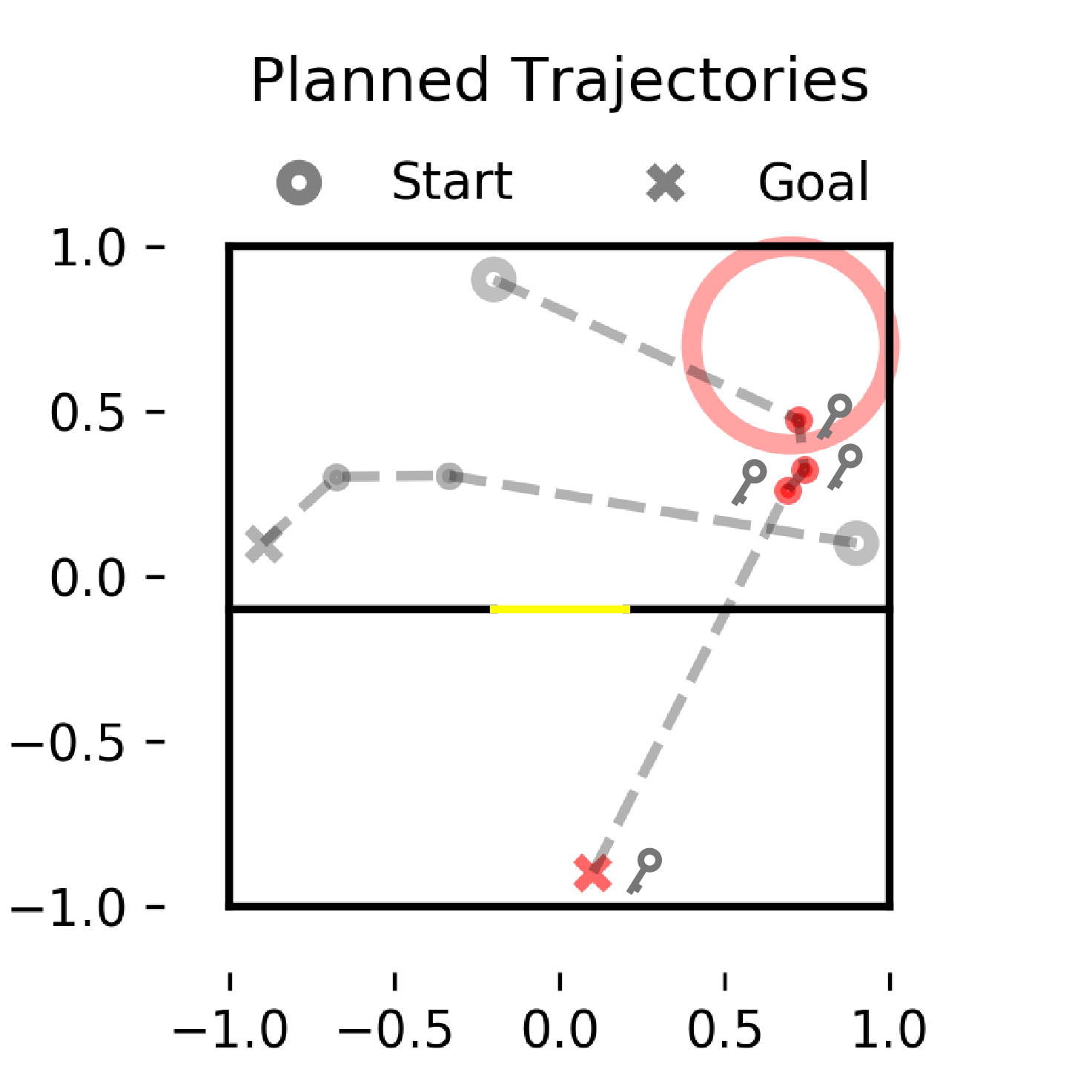}
    }
    \caption{}
  \end{subfigure}
    \caption{
    2D particle results on the $\epsilon$-key domain where the key dimension is scaled down to 0.1. 
    (a) \textbf{The key domain}: The rooms are separated in-between by a wall with a door (yellow). The door only opens when the agent has the key, which can be obtained if the agent is within the area indicated by the red circle on the upper right corner.
    (b) From no-key to has key, \textbf{k-means}: Value indicates the probability for the agent to transition from a state not having the key to a state having the key at each $(x, y)$ location. This transition should only occur near the key region (indicated by the red ring). 
    In this case, K-means fails to learn the separation between having and not having the key, and generated high transition probability over the entire domain.
    (c) The same figure as (b), generated by the \textbf{Causal InfoGAN}. On the top right corner where the key is located, the GAN correctly learns that it can transition from having no key to having the key. Bottom blots appear where the posterior sees no data. 
    (d) Causal-InfoGAN planned walkthrough trajectories, showing how the agent acquires the key to cross the door. When the goal is in the top room, the agent goes directly towards the goal without making a detour for the key region.
    }
    \label{fig:toy_task_key}
\end{figure}

Our toy domains involve a particle moving in a 2-dimensional continuous domain with impenetrable obstacles, as depicted in Figures \ref{fig:toy_task} and \ref{fig:toy_task_key}. The observations are the $(x,y)$ coordinates of the particle in the plane, and, in the door-key domain, also a binary indicator for holding the key. We generate data trajectories by simulating a random motion of the particle, started from random initial points. We consider the following various geometrical arrangements of the domain, chosen to demonstrate the properties of our method.
\begin{enumerate}%[leftmargin=2em]
    % \item \textbf{Rectangle:}
    \item \noindent \textbf{Tunnels:} the domain is partitioned into two unconnected rooms (top/bottom), where in each room there is an obstacle, positioned such that transitioning between the left/right quadrants is through a narrow tunnel. 
    \item \textbf{Door-key:} two rooms are connected by a door. The door can be traversed only if the agent holds the key, which is obtained by moving to the red-marked area in the top right corner of the upper room. Holding the key is represented as a binary 0/1 element in the observation. 
    \item \textbf{Rescaled door-key:} Same as door key domain, but the key observation is rescaled to be a small $\epsilon$ when the agent is holding the key, and 0 otherwise.
\end{enumerate}

Our domains are designed to distinguish when standard state aggregation methods, which rely on the Euclidean metric, can work well.
In the tunnel domain, the Euclidean metric is not informative about the dynamics in the task -- two points in different rooms inside the tunnel can be very close in Euclidean distance, but not connected, while points in the same room can be more distant but connected. In the door-key domain, the Euclidean distance is informative if observations with key and without key are very distant in Euclidean space, as in the 0/1 representation (compared to the domain size which is in $[-1,1]$). In the rescaled door-key, we make the Euclidean distance less informative by changing the key observation to be 0/$\epsilon$.

We compare Causal InfoGAN with several recent methods for aggregating observation features into states for planning. Note that in these simple 2D domains, feature extraction is not necessary as the observations are already low dimensional vectors. The simplest baseline is K-means, which relies on the Euclidean distance between observations. In \cite{baram2016spatio}, a variant of K-means  for temporal data was proposed, using a window of consecutive observations to measure a smoothed Euclidean distance to a cluster centroids. We refer to this method as temporal K-means. In \cite{mahadevan2007proto}, and more recently \cite{lakshminarayanan2016option} and \cite{machado2017laplacian}, spectral clustering (SC) was proposed to learn connected clusters in the data. For continuous observations, SC requires a distance function to build a connectivity graph, and previous studies \cite{mahadevan2007proto,lakshminarayanan2016option,machado2017laplacian} relied on the Euclidean distance, either by using nearest neighbor to connect the graph, or by using the exponentiated distance to assign edge weights.

In Figure \ref{fig:toy_task}, we show the Causal InfoGAN classification of observations to abstract states, $Q(s|o)$, and compare with the K-means baseline; the other baselines gave qualitatively similar results. Note that Causal InfoGAN learned a clustering that is related to the dynamical properties of the domain, while the baselines, which rely on a Euclidean distance, learned clusters that are not informative about the real possible transitions. As a result, the Causal InfoGAN clearly separates abstract states within each room, while the K-means baseline clusters observations across the wall. This demonstrates the potential of Causal InfoGAN to learn meaningful state abstractions without requiring a distance function in observation space. In Figure \ref{fig:toy_task_key} we show similar results for the door-key domain. When the key observation was scaled to $0/\epsilon$, standard clustering methods did not separate states with the key and without key to different clusters. Causal InfoGAN, on the other hand, learned a binary predicate for holding the key, and learned that obtaining the key happens in the correct position.

% \TODO{This paragraph needs some more work or add ref to appendix.}
To evaluate planning performance, we hand-coded an oracle function that evaluates whether an observation trajectory is feasible or not (e.g., does not cross obstacles, correctly reports $\emptyset$ when a trajectory does not exist). For causal InfoGAN, we ran the planning algorithm described in Section \ref{s:planning}. For baselines, we calculated cluster transitions from the data, and generated planning trajectories in observation space by using the cluster centroids. We chose algorithm parameters and stopping criteria by measuring the average feasibility score on a validation set of start/goal observations, and report the average feasibility on a held out test set of start/goal observations. We report our results in Table \ref{table:toy_tasks}. The Causal InfoGAN learned clusters that respect the causal properties of the domain, resulting in significantly better planning.
% Note that Causal InfoGAN significantly outperforms baseline methods on the tunnels and rescaled key domain -- in which the Euclidean distance is not informative for planning. 

% Raw data here: 
% door @ key scale 0.1 with kmeans dev: 0.00%; test: 24.50%
% door @ key scale 0.1 with temporal-kmeans dev: 0.00%; test: 17.50%
% door @ key scale 0.1 with spectral-clustering dev: 20.00%; test: 31.50%
% door @ key scale 1.0 with kmeans dev: 100.00%; test: 56.58%
% door @ key scale 1.0 with temporal-kmeans dev: 100.00%; test: 56.58%
% door @ key scale 1.0 with spectral-clustering dev: 60.00%; test: 56.58%
% tunnel @ key scale 1.0 with kmeans dev: 16.62%; test: 12.25%
% tunnel @ key scale 1.0 with temporal-kmeans dev: 18.38%; test: 7.00%
% tunnel @ key scale 1.0 with spectral-clustering dev: 18.81%; test: 8.75%

% GY: Aviv, the dev vs test set on the tunnel domain seems to differ quite a bit.

% \begin{table}[]
%     \centering
%     \begin{tabular}{|c|c|c|c|}
%     \hline
%                         & Tunnels & Door-key & Rescaled door-key \\ \hline
%     Causal-InfoGAN      &  98\%   &          &                   \\ \hline
%     K-means             & 12.25\% &  100\%   & 0.0\%             \\ \hline
%     Temporal K-means    &   7.0\% &  100\%   & 0.0\%             \\ \hline
%     Spectral clustering &  8.75\% &   60\%   & 20.0\%            \\ \hline
%     \end{tabular}
    
%     \caption{Planning results for illustrative tasks. Table shows average feasibility of plans (higher is better) generated by the different algorithms. Note that Causal-InfoGAN siginificantly outperforms baselines in domains where the Euclidean distance is not informative for planning.}
%     \label{table:toy_tasks}
% \end{table}

\begin{table}[h]
    \centering
    \begin{tabular}{|c|c|c|c|}
    \hline
                        & Tunnels & Door-key & Rescaled door-key \\ \hline
    Causal-InfoGAN      &    \textbf{98\%} &   98\%   &     \textbf{97\%}          \\ \hline
    % InfoGAN             &         &          &                   \\ \hline
    K-means             & 12.25\% &  \textbf{100\%}   &    0.0\%          \\ \hline
    Temporal K-means    &   7.0\% &  \textbf{100\%}   &    0.0\%          \\ \hline
    Spectral clustering &  8.75\% &   60\%   &   20.0\%          \\ \hline
    \end{tabular}
    \vspace{1em}
    \caption{Planning results for illustrative 2D tasks. Table shows average feasibility of plans (higher is better) generated by the different algorithms. Note that Causal-InfoGAN significantly outperforms baselines in domains where the Euclidean distance is not informative for planning.}
    \label{table:toy_tasks}
\end{table}

% \vspace{-1em}
\subsection{Rope Manipulation} \label{sec:rope-manipulation}
% \vspace{-0.5em}

In this section we demonstrate Causal InfoGAN on the task of generating realistic robotic rope manipulation trajectories from start to goal configurations. Then, we show that Causal InfoGAN generates significantly better trajectories than those generated by the state-of-the-art generative model baselines both visually and numerically.

\begin{figure}[h!]
    \centering
    \begin{subfigure}[b]{0.90\linewidth}
    \includegraphics[width=\textwidth]{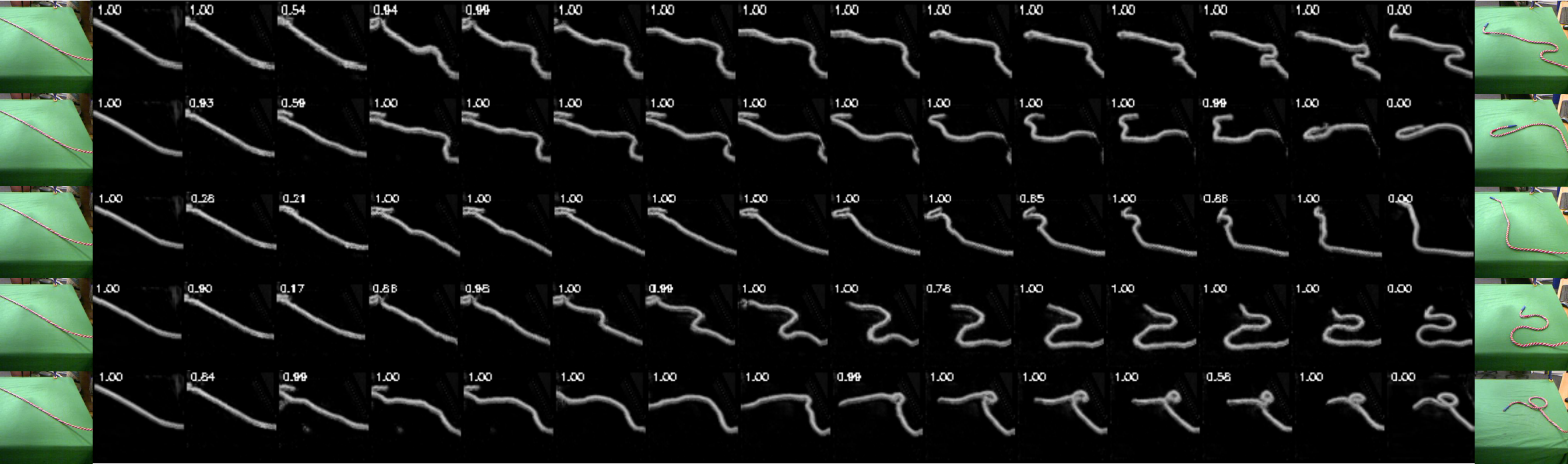}
    \caption{Causal InfoGAN}
    \label{fig:rope_cigan}
    \end{subfigure}\hfill
    
    \begin{subfigure}[b]{0.90\linewidth}
    \includegraphics[width=\textwidth]{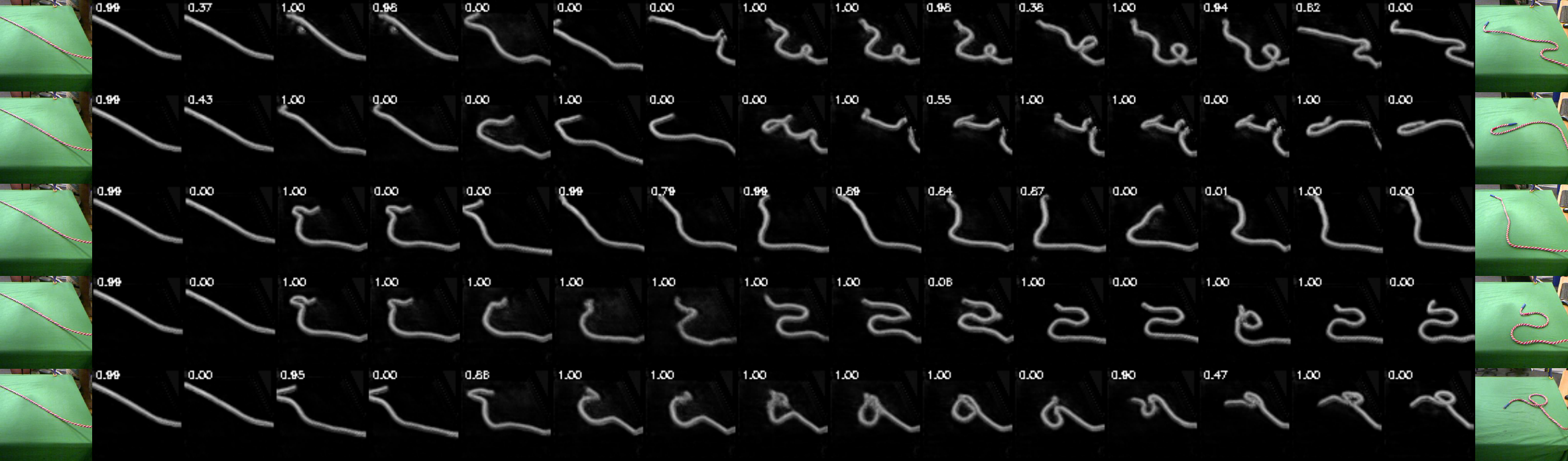}
    \caption{InfoGAN}
    \label{fig:rope_igan}
    \end{subfigure}
    
    \begin{subfigure}[b]{0.90\linewidth}
    \includegraphics[width=\textwidth]{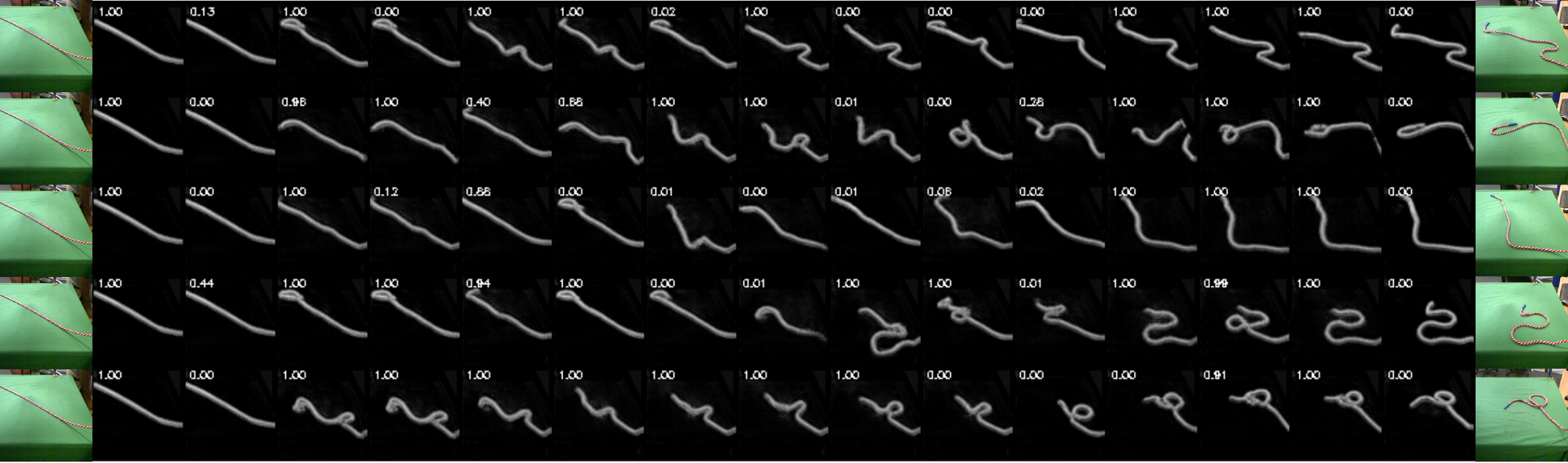}
    \caption{DCGAN}
    \label{fig:rope_dcgan}
    \end{subfigure}
    
    \caption{Results for rope manipulation data. We compare planning using Causal InfoGAN (top), InfoGAN (middle), and DCGAN (bottom) by interpolation in the latent space, for several rope manipulation goals starting from the same initial configuration. Each plot shows 5 planning instances, from left (starting observation) to the right (goal observation). For each instance, the shown trajectory is picked using the highest trajectory score. The training loss of Causal InfoGAN led to a latent space that is the most accurately represents possible changes to the rope, compared to the other two baselines.}
    \label{fig:rope}
    % \vspace{-1.5em}
\end{figure}

\begin{SCfigure}%[h!]
    % \centering
    \caption{Evaluation of walkthrough planning in rope domain. We trained a classifier to predict whether two observations are sequential or not (1=sequential, 0=not sequential), and compare the average classification score for different generative models. Note that Causal InfoGAN significantly outperforms the baselines, in alignment with the qualitative results of Figure \ref{fig:rope}.}%\DONE{made fonts larger}}
    \includegraphics[width=0.37\textwidth]{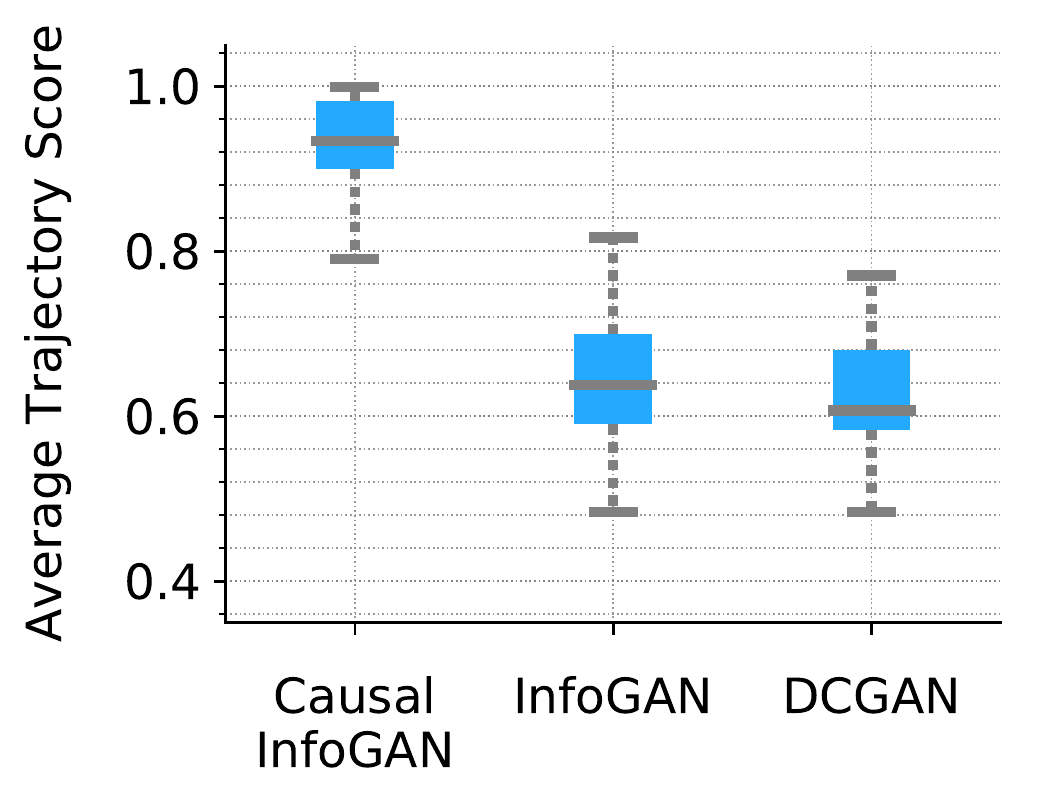}
    \label{fig:rope-boxplot}
\end{SCfigure}

The rope manipulation dataset~\cite{nair2017combining} contains a set of sequential images of a robot manipulating a rope in a self-supervised manner, by randomly choosing a point on the rope and perturbing it slightly. Using this data, the task is to manipulate the rope in a goal-oriented fashion, from one configuration to another, where a goal is represented as an image of the desired rope configuration. In the original study, Nair et al.~\cite{nair2017combining} used the data to learn an inverse dynamics model for manipulating the rope between two images of similar rope configurations. Then, to solve long-horizon planning, Nair et al.~required a human to provide the walkthrough sequence of rope poses, and used the learned controller to execute the short-horizon transitions within the plan. 

% In a study on self-supervised robot learning, Nair et al.~\cite{nair2017combining} generated a dataset of sequential images of a robot manipulating a rope in a self-supervised manner, by randomly choosing a point on the rope and perturbing it slightly. 

In our experiment, we show that Causal InfoGAN can be used to generate walkthrough plans directly from data for long-horizon tasks, without requiring additional human guidance. We train a Causal-InfoGAN model on the rope manipulation data of \cite{nair2017combining}.
We pre-processed the data by removing the background, and applying a grayscale transformation. We chose the continuous abstract state representation described in Section \ref{s:planning}.
In Figure \ref{fig:rope_cigan}, we show our results for planning walkthroughs between different rope observations. Note that planning here is simply interpolation in the abstract space, however, the Causal InfoGAN objective guarantees that such interpolation indeed relates to feasible transitions. In comparison, in Figure \ref{fig:rope_igan}, we trained a standard InfoGAN model, where the mutual information loss does not involve state transitions, and perform interpolation in the InfoGAN abstract state space. We also trained a standard DCGAN model as another baseline, where the observations do not share mutual information with the abstract states, as shown in in Figure \ref{fig:rope_dcgan}.\footnote{For DCGAN and InfoGAN, Encoding an observation to a latent state, and decoding a latent state trajectory to an observation walkthrough were done using a similar approach to the Causal InfoGAN method described in Section \ref{s:planning}.} We see that, due to the causality preserving loss, only Causal InfoGAN learns a smooth latent space in which linear interpolation indeed correspond to plausible trajectories in the observation space.

Unlike the synthetic 2D domains above, numerically evaluating planning performance is difficult, since we cannot design a perfect oracle for evaluating the feasibility of a generated visual plan. Instead, we propose a surrogate evaluation criteria: we exploit the fact that we have data of full manipulation trajectories, and train a binary classifier to classify whether two images are sequential in the data or not\footnote{The positive data are the pairs of rope images that are 1 step apart and the negative data are randomly chosen pairs that are from different runs which are highly likely to be farther than 1 step apart.}. For an image pair, the classifier output therefore provides a score between 0 and 1 for the feasibility of the transition.
% give the closeness score of image pairs between 0 (far) and 1 (close). 
 We apply the classifier to compute the \textit{trajectory score} which is the average classifier score of image pairs in the trajectory. Note that this classifier is trained independent of the generative models. Thus, the trajectory score is an impartial metric. For each start and goal, we pick the best trajectory score out of 400 samples of the noise variable $z$.\footnote{This selection process is applied the same way to the DCGAN and InfoGAN baselines.} As shown in Figure \ref{fig:rope-boxplot}, Causal InfoGAN achieved a significantly higher trajectory score averaged over 57 task configurations.

\section{Conclusion}
We presented Causal InfoGAN, a framework for learning deep generative models of sequential data with a structured latent space. 
By choosing the latent space to be compatible with efficient planning algorithms, we developed a framework capable of generating goal-directed trajectories from high-dimensional dynamical systems. 

Our results for generating realistic manipulation plans of a rope suggest promising applications in robotics, where designing models and controllers for manipulating deformable objects is challenging. 

The binary latent models we explored provide a connection between deep representation learning and classical AI planning, where Causal InfoGAN can be seen as a method for learning object predicates directly from data. In future work we intend to investigate this direction further, and incorporate object-oriented models, which are a fundamental component in classical AI.

% state representations that (1) capture the dynamics in the data; and (2) allow for efficient long horizon planning. We showed that our method can be used to learn from unstructured exploration data how to generate a `walk-through' of the system transitioning between a pair of start and goal observations. Such a walk-through can be later used to guide a low-level controller in executing the task.

% \TODO{Add outlook}

\newpage
\bibliography{references}
\bibliographystyle{abbrv}

\newpage

\appendix

\section{Algorithm}\label{sec:algorithm}

Given the training data $\data = \left\{(o,o')|\mbox{ $o$ and $o'$ are sequential observations.}\right\}$, Causal infoGAN learns a generative model that structures the latent space in a way that is useful for planning. We provide the algorithm details below:

Let $\theta_D, \theta_G, \theta_Q, \theta_T$ denote the parameters of neural networks $D,G,Q,T_{\model}$, respectively.

For a minibatch of $m$ samples $\left\{(o_{i},o'_{i})\right\}_{i=1}^m$ from $\data$, we:
\begin{itemize}
    \item Generate $m$ fake samples
    \begin{itemize}
        \item Sample abstract states $s_{1},\dots,s_{m}$, where $s_{i} \sim P_{\model}$
        \item Sample next states $s'_{1},\dots,s'_{m}$, where $s'_{i} \sim T_{\model}(s'_{i}|s^{i})$
        \item Sample noise $z_{1},\dots,z_{m}$, where $z_{i}\sim \PN$
        \item Generate fake observations $\hat{o}_{1}, \hat{o}'_{1},\dots, \hat{o}_{m}, \hat{o}'_{m}$, where $\hat{o}_{i}, \hat{o}'_{i} = G(z_{i},s_{i},s'_{i}). $
        % \item Compute most likely state encoding $s^{*,1},s'^{*,1}\dots,s^{*,m},s'^{*,m}$ for real data: $s^{*,i} = \argmax_s Q(s|o^{i})$, $s'^{*,i} = \argmax_s Q(s|o'^{i})$.

    \end{itemize}
    \item Update the discriminator by descending its stochastic gradient
    \begin{equation*}
        \nabla_{\theta_D} \left(-\frac{1}{m}\sum_{i=1}^{m}\left[ \log D(o_{i}, o'_{i}) + \log (1 - D(\hat{o}_{i}, \hat{o}'_{i}))\right]\right)
    \end{equation*}
    \item Update the generator and transition model by descending its stochastic gradient
    \begin{equation*}
        \nabla_{\theta_G,\theta_T} \left(-\frac{1}{m}\sum_{i=1}^{m}\left[ \log D(\hat{o}_{i}, \hat{o}'_{i}) \right]\right),
    \end{equation*}
    where the gradient $\theta_T$ is backpropagated using the reparametrization trick of Gumbel-softmax~\cite{jang2016categorical}.
    \item Update posterior, generator, and transition model in the direction of maximal mutual information
    \begin{equation*}
        \nabla_{\theta_Q,\theta_G,\theta_T} \left(\frac{1}{m}\sum_{i=1}^{m}\left[ \log P_{\model}(s_{i}) -\log Q(\hat{o}_{i} | s_{i}) + \log T_{\model}(s'_{i} | s_{i}) -\log Q(\hat{o}'_{i} | s'_{i})  \right]\right),
    \end{equation*}
    where the gradient $\theta_T$ is backpropagated using the reparametrization trick of Gumbel-softmax~\cite{jang2016categorical}.
    \item (\textit{For continuous states with linear interpolation planning}) Update transition model to ensure small local transitions in the state space generate plausible observations.
    \begin{equation*}
        \nabla_{\theta_T} \left(\frac{1}{m}\sum_{i=1}^{m} \norm{ \Sigma_{\theta_T}(s_i) }_2  \right),
    \end{equation*}
    where $\Sigma_{\theta_T}$ is part of $T_\model$ (see Section \ref{ss:latent_systems}).
    
    \item (\textit{Optional}) Update transition model to minimize a self-consistency loss (see details below) 
    \begin{equation*}
        \nabla_{\theta_T} \left(\frac{1}{m}\sum_{i=1}^{m}\left[ -\log T_{\model}(s^*(o'_i) | s^*(o_i))\right]\right),
    \end{equation*}
    % where $\textrm{CE}(x,y)$ denotes the average of the element-wise cross-entropy loss between the elements of $x$ and $y$.
    where $s^*(o) = \argmax_s Q(s|o)$.
\end{itemize}

The self-consistency loss is added to further strengthen the relationship between transitions in the latent planning system and the real observations. We maximize the likelihood of observed transitions in the predicted states from real transitions. Namely, let $s^*(o) = \argmax_s Q(s|o)$ denote the most likely state encoding for an observation, then the self-consistency loss is given by, 
%% TK: In theory, we don't need argmax and can use the distribution given by Q(s|o) to compute self-consistency loss.
%% AT: That's true, but for sake of clarity, not sure we want to delve into it here. Do you think it is important to add?
\begin{equation*}
    L_{sc}(\model) = \mathbb{E}_{o,o'\sim \PD} \left[ -\log T_{\model}(s^*(o') | s^*(o))\right].
\end{equation*}
This loss guides $T$ to be consistent with $Q$ which stabilizes the training. We found this loss to help in stabilizing training for low-dimensional observations. We did not find that adding this loss is beneficial in the high-dimensional case, since in that case, while $Q$ provides meaningful state estimation on fake observations, it tends to overfit to the generated samples, and does not predict reliable states on 
% collapse to a single state given 
real observations. 
% This means $s^*(o')$ are $s^*(o)$ close (continuous) or the same (discrete). Thus, the self-consistency loss encourages $T$ to be stationary, which acts similarly to the $L_{cont}$ in Equation \ref{eq:lcont}. 

\section{Experiment details}\label{sec:experiment_details}

\subsection{2D Navigation Experiment}

The model parameters we used for the toy domains is as follows:

In the key domain, we used a 4-dimensional space for the latent state (we also experimented with 3-5 dimensional latent spaces which gave similar results. Smaller latent space tend to have less expressive power and subject to generator collapse. Beyond 5 however the benefit is marginal.) Actions are sampled from a 3 dimensional space whereas the noise $z$ is 4 dimensional. The loss for the generator, the posterior and the transition consistency are weighted equally with a learning rate of $10^{-4}$, whereas the learning rate for the discriminator is five times larger ($5\times 10^{-4}$). We found that the transition consistency loss important in the stability of models using binary representations. The same hyperparameters are used for both the key domain and the $\epsilon$-key domain. In the latter $\epsilon=0.1$.

In the tunnel domain we also used a 4-dimensional latent state (a 3-dimensional latent state gave similar results). Actions are sampled from a 3-dimensional space and noise is 4 dimensional. The learning rates are identical to those of the key domain. 

To generate the training samples in the tunnel domain, the random walk had a characteristic length scale of 0.05. 
% We take step-skip of 5, meaning data points are sampled 5 timesteps apart. 
The rooms are from -1 to 1 in both width and height. The meridian is placed slightly off the middle, at y==-0.1. We bias the starting point in the particle trajectories around the choke in the middle, so that the sample trajectories have substantial probability is crossing from one room to the other.

In the key domain, we used a a characteristic length scale of 0.3.
% in both x and y (non-Cartesian) with a step-skip size between 4 and 9 (inclusive). 
This much larger step size is needed because the particle needs to cover the top room, make it to the key zone (to obtain the key), and carry the key to the door to cross to the bottom room in a single trajectory.

In the tunnel domain, we chose the horizon $k$ to be uniform in $5-9$. In the key domain, since the charecteristic length scale is larger, we chose $k$ to be in $1-4$.

% During our experiment, we found that a large step step (4 - 8) also helps in stabilizing the training. 
In the key domain, we represent the possession of the key by a single number in the binary set \{0, 1\}. Incidentally, it was necessary to inject Gaussian noise to this key dimension during training. Otherwise the generator is required to learn a singularity around 0 and 1, making it numerically highly unlikely. We varied the normalized standard deviation of this Gaussian noise (w.r.t. $\epsilon$). Larger noise ($\leq 0.2$) produces more stable training, but too much noise can cause blurriness in the cluster boundaries. Overall the scale of this Gaussian noise doesn't substantially impact the representation that is learned.

The models are identical between the key domain and the tunnel domain. Both the generator, the discriminator and the binary posterior are two layer perceptrons with two 100 dimensional hidden layers. The transition function also has two hidden layers, with 10 neurons each. 

For the represetantion of the latent space, we use a binary representation of the states as described in section \ref{s:binary_state}. The generator uses a sequential architecture as described in Section \ref{ss:decoding_trajectory}, but the two outputs of the generator are trained with only 1 timestep in between with no autoregression on the autoregressive sub module.

\subsection{Rope Experiment}

We use Adam \cite{kingma2014adam} optimizer with the learning rate of 0.0002 for both discriminator and generator losses. The generator loss is the sum of three losses described in the Appendix \ref{sec:algorithm}. We use coefficients 1 for the main generator loss, and 0.1 for both the mutual information and the transition loss. We deploy standard DCGAN architectures \cite{radford2015unsupervised} for the discriminator D and the generator G. The posterior estimator Q has the same architecture as D with the change of the last CNN layer to output 128 channels and the addition of another layer of batchnorm, leaky ReLU and conv layer to the dimension of code. The details are described in table \ref{tab:architecture}. 

In DCGAN and infoGAN baselines, the size of latent code (or abstract state) and the noise is 7 and 2 respectively. In Causal InfoGAN, the generator takes in two abstract states at the same time so the size of noise is doubled to 4. However, we found that the result is quite robust to the dimension sizes. 

\begin{table}[hb!]

\begin{tabular}{|c | c|} 
\hline
discriminator D / posterior estimator Q & generator G \\ [0.5ex] 
\hline\hline
Input 64 x 64 grayscale images
(2 for D and 1 for Q)  & Input a vector in $\mathbf{R}^{7} \times \mathbf{R}^{7} \times \mathbf{R}^{4}$ \\ 
\hline
4 x 4 conv. 64 lReLU, stride 2 , batchnorm  & 4 x4 upconv. 512 lReLU, stride 2 , batchnorm  \\
\hline
4 x 4 conv. 128 lReLU, stride 2 , batchnorm  & 4 x4 upconv. 256 lReLU, stride 2, batchnorm \\
\hline
4 x 4 conv. 256 lReLU, stride 2 , batchnorm  & 4 x4 upconv. 128 lReLU, stride 2, batchnorm \\
\hline
4 x 4 conv. 512 lReLU, stride 2 , batchnorm  & 4 x4 upconv. 64 lReLU, stride 2, batchnorm \\ 
\hline
4 x 4 conv. 1 for D and 128-batchnorm-lReLU-7 for Q & 4 x4 upconv. 2 Tanh (1 channel for each image) \\ [0.5 ex]
\hline 
\end{tabular} \\ [1 ex]
\caption{The architectures for generating rope images. The discriminator takes in two grayscale images, and outputs the probability of the pair being real. The posterior shares the same architecture with D except the first and the last layer. It takes in one image, and outputs the mean and the variance of its predicted state. The generator takes in the current and the next abstract states (dim 7), and the noise (dim 4). It outputs the current and the next observations. The leaky coefficient is 0.2.}
\label{tab:architecture}
\end{table}

The latent planning system uses a uniform prior between $[-1,1]$ and a Gaussian transition with zero mean and state-dependent variance. The variance is diagonal and parametrized by a two-layer feed forward neural network of size 64 with ReLU nonlinearity. The last layer is exponentiated to output a positive value for the variance. 

For training set, we use sequential observation pairs with 1 step apart from the rope dataset by Nair et al. \cite{nair2017combining}.
\subsubsection{Causal Classifier}

We trained a binary classifier to function as an evaluator for whether an observation transition is feasible or not, given the data.
We use the classifier for two tasks: (1) To post-select transitions (in the observation space) during planning, and (2) to evaluate the score of a walkthrough trajectory. 

During training, the classifier takes in a pair of images and output a binary classification of whether this image pair appears sequentially related. The training dataset consists of positive image pairs that are $1$ timestep apart, and negative pairs that are randomly sampled from different rope manipulation runs. To avoid overfitting to the background in the rope dataset and learning a trivial solution where the classifier uses the background to distinguishi different runs, we preprocess the rope data using the background subtraction pipeline mentioned above. 

The training accuracy converges to 100\% on the training set, and 98\% on a held-out test set. 

To validate that this classifier actually learns to tell if the transition between two images is feasible or not, we evaluate it on images that are $k$ steps apart where the largest k is the length of an rope experiment. Despite the classifier never seeing samples that are more than 1 step apart, it learns to predict $0$ probability for image pairs with large $k$. The prediction is well-behaved -- As we increase k from 1 to the length of the run, the binary output smoothly and monotonically decreases from $1$ to $0$.

The model architecture used is a convolutional neural network with the following architecture. The two input images are concatenated channel-wise, fed together into the classifier. The optimization is done with the Adam optimizer with a learning rate of $10^{-3}$. These hyperparameters are not tuned since the performance of the classifier is sufficient.

% \begingroup
%     \fontsize{8p}{10pt}\selectfont
%     \begin{verbatim}
%     nn.Sequential([
%         # input size (2 or 6) x 64 x64. Take 2 black and white images.
%         nn.Conv2d(2, 64, 4, 2, 1),
%         nn.LeakyReLU(0.1, inplace=True),
%         # 64 x 32 x 32
%         nn.Conv2d(64, 128, 4, 2, 1, bias=False),
%         nn.BatchNorm2d(128),
%         nn.LeakyReLU(0.1, inplace=True),
%         # 128 x 16 x 16
%         nn.Conv2d(128, 256, 4, 2, 1, bias=False),
%         nn.BatchNorm2d(256),
%         nn.LeakyReLU(0.1, inplace=True),
%         # Option 1: 256 x 8 x 8
%         nn.Conv2d(256, 512, 4, 2, 1, bias=False),
%         nn.BatchNorm2d(512),
%         nn.LeakyReLU(0.1, inplace=True),
%         # 512 x 4 x 4
%         nn.Conv2d(512, 1, 4),
%         h.Flatten(),
%         nn.Sigmoid()
%     ])
% \end{verbatim}
% \endgroup

\end{document}